\documentclass[10pt,twocolumn,letterpaper]{article}

\usepackage{cvpr}
\usepackage{times}
\usepackage{epsfig}
\usepackage{graphicx}
\usepackage{amsmath}
\usepackage{amssymb}
\usepackage{algorithm}
\usepackage{algpseudocode}
\usepackage[font=small]{caption}
\usepackage{subcaption}
\usepackage{xfrac}
\usepackage{microtype}
\usepackage{booktabs}
\usepackage{siunitx}

\AtBeginDocument{%
 \textfloatsep=10pt plus 2pt minus 4pt
 \floatsep=5pt plus 2pt minus 2pt
}

\makeatletter
\renewcommand{\ALG@beginalgorithmic}{\small}
\makeatother

\graphicspath{ {images/} }

\usepackage[pagebackref=true,breaklinks=true,letterpaper=true,colorlinks,bookmarks=false]{hyperref}

\cvprfinalcopy 

\def\cvprPaperID{360} 

\begin{document}

\title{Spatially Adaptive Computation Time for Residual Networks}

\author{Michael Figurnov$^{1}$\thanks{This work was done while M. Figurnov was an intern at Google.} \quad Maxwell D. Collins$^2$ \quad Yukun Zhu$^2$ \quad Li Zhang$^2$ \quad Jonathan Huang$^2$ \\
Dmitry Vetrov$^{1,3}$ \quad Ruslan Salakhutdinov$^4$ \\
$^1$National Research University Higher School of Economics \quad $^2$Google Inc. \\
$^3$Yandex \quad $^4$Carnegie Mellon University \\
{\tt \small michael@figurnov.ru \quad \{maxwellcollins,yukun,zhl,jonathanhuang\}@google.com} \\
{\tt \small vetrovd@yandex.ru \quad rsalakhu@cs.cmu.edu }
}

\maketitle

\begin{abstract}
This paper proposes a deep learning architecture based on Residual Network that dynamically adjusts the number of executed layers for the regions of the image.
This architecture is end-to-end trainable, deterministic and problem-agnostic.
It is therefore applicable without any modifications to a wide range of computer vision problems such as image classification, object detection and image segmentation.
We present experimental results showing that this model improves the computational efficiency of Residual Networks on the challenging ImageNet classification and COCO object detection datasets.
Additionally, we evaluate the computation time maps on the visual saliency dataset cat2000 and find that they correlate surprisingly well with human eye fixation positions.
\end{abstract}

\section{Introduction}
Deep convolutional networks gained a wide adoption in the image classification problem~\cite{krizhevsky2012imagenet,simonyan2014verydeep,szegedy2015inception} due to their exceptional accuracy.
In recent years deep convolutional networks have become an integral part of  state-of-the-art systems for a diverse set of computer vision problems such as object detection~\cite{ren2015faster}, image segmentation~\cite{long2015fully}, image-to-text~\cite{karpathy2015deep,xu2015show}, visual question answering~\cite{fukui2016multimodal} and image generation~\cite{dosovitskiy2015learning}.
They have also been shown to be surprisingly effective in non-vision domains, \eg natural language processing~\cite{zhang2015character} and analyzing the board in the game of Go~\cite{silver2016mastering}.

A major drawback of deep convolutional networks is their huge computational cost.
A natural way to tackle this issue is by using attention to guide the computation, which is similar to how biological vision systems operate~\cite{rensink2000dynamic}.
Glimpse-based attention models~\cite{larochelle2010learning,mnih2014recurrent,ba2015multiple,jaderberg2015spatial} assume that the problem at hand can be solved by carefully processing a small number of typically rectangular sub-regions of the image.
This makes such models unsuitable for multi-output problems (generating box proposals in object detection) and per-pixel prediction problems (image segmentation, image generation).
Additionally, choosing the glimpse positions requires designing a separate prediction network or a heuristic procedure~\cite{almahairi2016dynamic}.
On the other hand, soft spatial attention models~\cite{xu2015show,sharma2016action} do not allow to save computation since they require evaluating the model at all spatial positions to choose per-position attention weights.

\begin{figure}
    \centering
    \begin{subfigure}{\linewidth}
        \centering
        \includegraphics[width=0.46\linewidth]{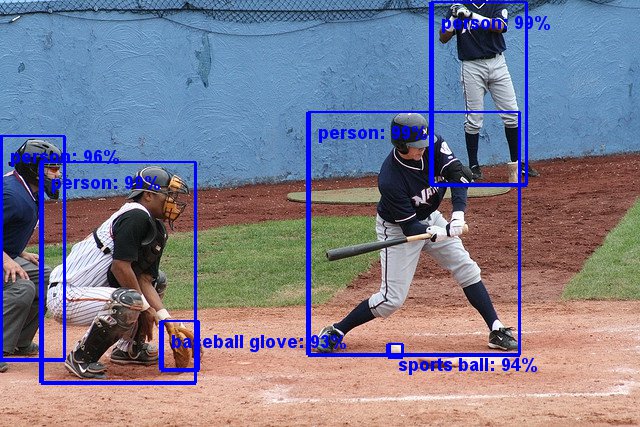}
        \includegraphics[width=0.46\linewidth]{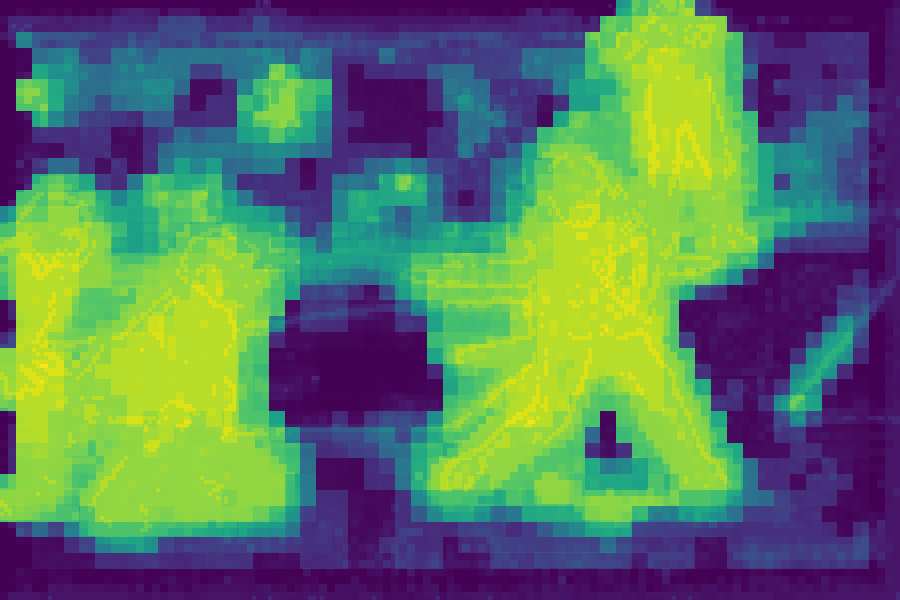}
        \includegraphics[width=0.04825\linewidth]{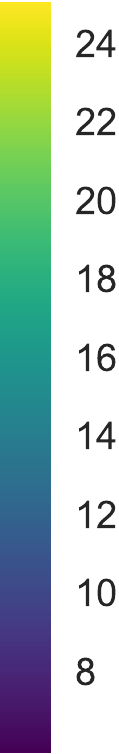}
    \end{subfigure}
    \caption{Left: object detections. Right: feature extractor SACT ponder cost (computation time) map for a COCO validation image. The proposed method learns to allocate more computation for the object-like regions of the image.}
    \label{fig:teaser}
\end{figure}

We build upon the Adaptive Computation Time (ACT)~\cite{graves2016adaptive} mechanism which was recently proposed for Recurrent Neural Networks (RNNs).
We show that ACT can be applied to dynamically choose the number of evaluated layers in Residual Network~\cite{he2016deep,he2016identity} (the similarity between Residual Networks and RNNs was explored in ~\cite{liao2016bridging,greff2017highway}).
Next, we propose Spatially Adaptive Computation Time (SACT) which adapts the amount of computation between spatial positions.
While we use SACT mechanism for Residual Networks, it can potentially be used for convolutional LSTM~\cite{xingjian2015convolutional} models for video processing~\cite{li2016videolstm}.

SACT is an end-to-end trainable architecture that incorporates attention into Residual Networks.
It learns a deterministic policy that stops computation in a spatial position as soon as the features become ``good enough''.
Since SACT maintains the alignment between the image and the feature maps, it is well-suited for a wide range of computer vision problems, including multi-output and per-pixel prediction problems.

We evaluate the proposed models on the ImageNet classification problem~\cite{deng2009imagenet} and find that SACT outperforms both ACT and non-adaptive baselines.
Then, we use SACT as a feature extractor in the Faster R-CNN object detection pipeline~\cite{ren2015faster} and demonstrate results on the challenging COCO dataset~\cite{lin2014microsoft}. 
Example detections and a ponder cost (computation time) map are presented in fig.~\ref{fig:teaser}.
SACT achieves significantly superior FLOPs-quality trade-off to the non-adaptive ResNet model.
Finally, we demonstrate that the obtained computation time maps are well-correlated with human eye fixations positions, suggesting that a reasonable attention model arises in the model automatically without any explicit supervision.

\section{Method}

We begin by outlining the recently proposed deep convolutional model Residual Network (ResNet)~\cite{he2016deep,he2016identity}.
Then, we present Adaptive Computation Time, a model which adaptively chooses the number of residual units in ResNet.
Finally, we show how this idea can be applied at the spatial position level to obtain Spatially Adaptive Computation Time model.

\subsection{Residual Network}
\begin{figure}
    \centering
    \includegraphics[width=0.9\linewidth]{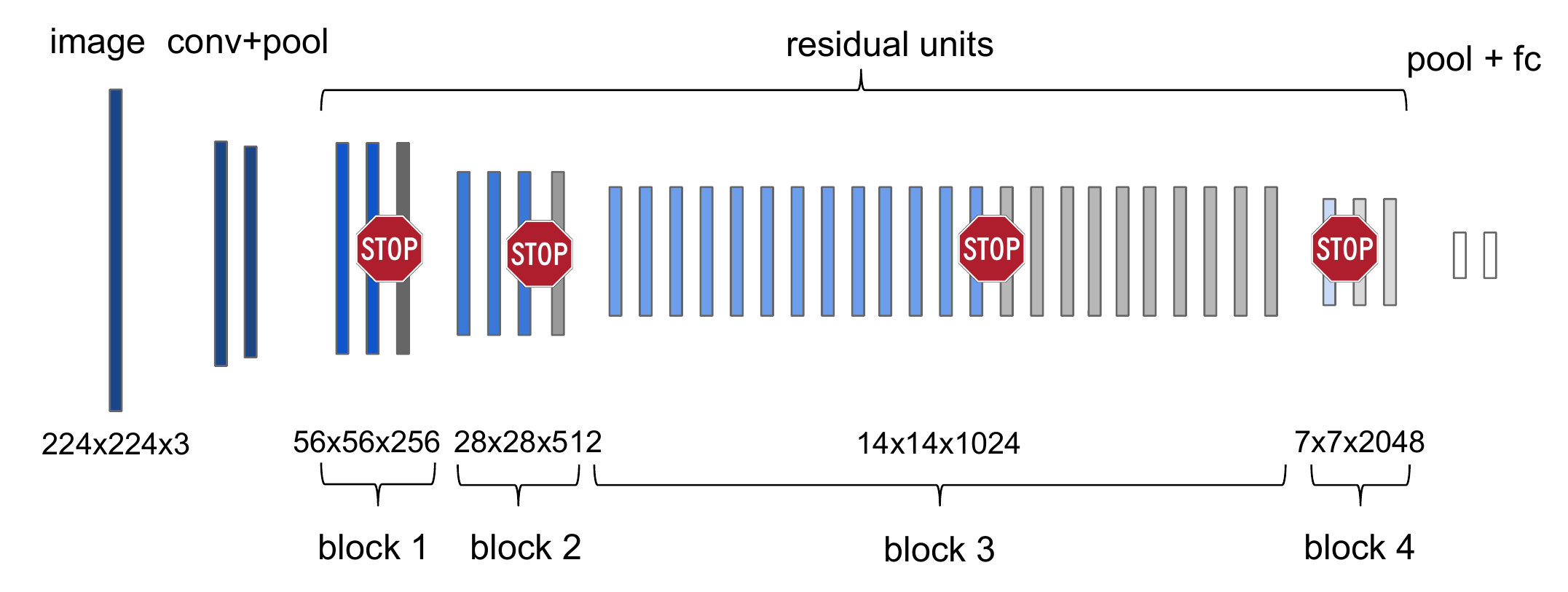}
    \caption{Residual Network (ResNet) with 101 convolutional layers. Each residual unit contains three convolutional layers. We apply Adaptive Computation Time to each block of ResNet to learn an image-dependent policy of stopping the computation.}
    \label{fig:resnet}
\end{figure}

We first describe the ResNet-101 ImageNet classification architecture (fig.~\ref{fig:resnet}).
It has been extended for object detection~\cite{he2016deep,dai2016rfcn} and image segmentation~\cite{chen2016deeplab} problems.
The models we propose are general and can be applied to any ResNet architecture.
The first two layers of ResNet-101 are a convolution and a max-pooling layer which together have a total stride of four.
Then, a sequence of four blocks is stacked together, each block consisting of multiple stacked \textit{residual units}.
ResNet-101 contains four blocks with 3, 4, 23 and 3 units, respectively.
A residual unit has a form $F(\mathbf{x}) = \mathbf{x} + f(\mathbf{x})$, where the first term is called a shortcut connection and the second term is a residual function.
A residual function consists of three convolutional layers: $1 \times 1$ layer that reduces the number of channels, $3 \times 3$ layer that has equal number of input and output channels and $1 \times 1$ layer that restores the number of channels.
We use \textit{pre-activation} ResNet~\cite{he2016identity} in which each convolutional layer is preceded by batch normalization~\cite{ioffe2015batch} and ReLU non-linearity.
The first units in blocks 2-4 have a stride of 2 and increases the number of output channels by a factor of 2.
All other units have equal input and output dimensions.
This design choice follows Very Deep Networks~\cite{simonyan2014verydeep} and ensures that all units in the network have an equal computational cost (except for the first units of blocks 2-4 having a slightly higher cost).

Finally, the obtained feature map is passed through a global average pooling layer~\cite{lin2014network} and a fully-connected layer that outputs the logits of class probabilities.
The global average pooling ensures that the network is \textit{fully convolutional} meaning that it can be applied to images of varying resolutions without changing the network's parameters.

\subsection{Adaptive Computation Time}

\begin{figure}
    \centering
    \includegraphics[width=0.9\linewidth]{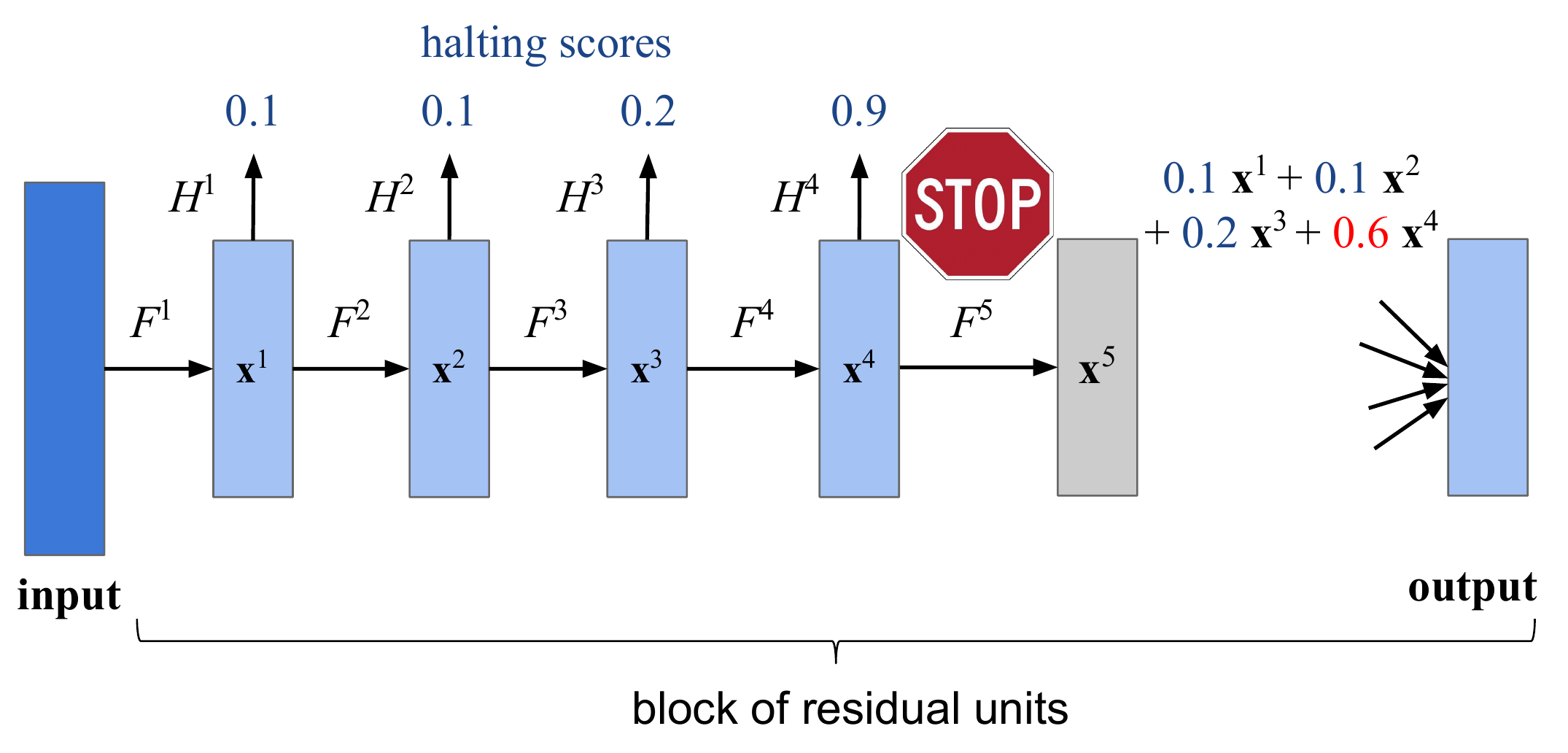}
    \caption{Adaptive Computation Time (ACT) for one block of residual units. The computation halts as soon as the cumulative sum of the halting score reaches 1. The remainder is $R = 1 - h^1 - h^2 - h^3 = \color{red}{0.6}$, the number of evaluated units $N = 4$, and the ponder cost is $\rho = N + R = 4.6$. See alg.~\ref{algorithm:act}. ACT provides a deterministic and end-to-end learnable policy of choosing the amount of computation.}
    \label{fig:act}
\end{figure}

Let us first informally explain Adaptive Computation Time (ACT) before describing it in more detail and providing an algorithm.
We add a branch to the outputs of each residual unit which predicts a \textit{halting score}, a scalar value in the range $[0, 1]$.
The residual units and the halting scores are evaluated sequentially, as shown in fig.~\ref{fig:act}. 
As soon as the cumulative sum of the halting score reaches one, all following residual units in this block will be skipped.
We set the halting distribution to be the evaluated halting scores with the last value replaced by a \textit{remainder}.
This ensures that the distribution over the values of the halting scores sums to one. 
The output of the block is then re-defined as a weighted sum of the outputs of residual units, where the weight of each unit is given by the corresponding probability value.
Finally, a \textit{ponder cost} is introduced that is the number of evaluated residual units plus the remainder value.
Minimizing the ponder cost increases the halting scores of the non-last residual units making it more likely that the computation would stop earlier.
The ponder cost is then multiplied by a constant $\tau$ and added to the original loss function.
ACT is applied to each block of ResNet independently with the ponder costs summed.

Formally, we consider a block of $L$ residual units (boldface denotes tensors of shape Height $\times$ Width $\times$ Channels):
\begin{align} 
    &\mathbf{x}^0 = \textbf{input}, \\
    &\mathbf{x}^l = F^l(\mathbf{x}^{l-1}) = \mathbf{x}^{l-1} + f^l(\mathbf{x}^{l-1}),\ l = 1 \dots L \label{eqn:resnet-update}, \\
    &\textbf{output} = \mathbf{x}^L.
\end{align}
We introduce a halting score $h^l \in [0, 1]$ for each residual unit.
We define $h^L = 1$ to enforce stopping after the last unit.
\begin{align}
    &h^l = H^l(\mathbf{x}^l),\ l = 1 \dots (L-1), \\
    &h^L = 1.
\end{align}
We choose the halting score function to be a simple linear model on top of the pooled features:
\begin{equation}
    h^l = H^l(\mathbf{x}^l) = \sigma (W^l \operatorname{pool}(\mathbf{x}^l) + b^l),
\end{equation}
where $\operatorname{pool}$ is a global average pooling and $\sigma(t) = \frac{1}{1 + \exp(-t)}$.

Next, we determine $N$, the number of residual units to evaluate, as the index of the first unit where the cumulative halting score exceeds $1-\varepsilon$:
\begin{equation}
    N = \min \Big\{ n \in \{1 \dots L\}: \sum_{l=1}^{n} h^l \geq 1 - \varepsilon \Big\},
    \label{eqn:N}
\end{equation}
where $\varepsilon$ is a small constant (\eg, 0.01) that ensures that $N$ can be equal to 1 (the computation stops after the first unit) even though $h^1$ is an output of a sigmoid function meaning that $h^1 < 1$.

Additionally, we define the remainder $R$:
\begin{equation}
    R = 1 - \sum_{l=1}^{N-1} h^l.
\end{equation}
Due to the definition of $N$ in eqn. \eqref{eqn:N}, we have $0 \leq R \leq 1$.

We next transform the halting scores into a \textit{halting distribution}, which is a discrete distribution over the residual units.
Its property is that all the units starting from $(N+1)$-st have zero probability:
\begin{equation}
    p^l = \begin{cases}
            h^l &\text{if } l < N, \\
            R &\text{if } l = N, \\
            0 &\text{if } l > N.
            \end{cases}
\end{equation}

\begin{algorithm}[t]
\begin{algorithmic}[1]
    \Require 3D tensor \textbf{input}
    \Require number of residual units in the block $L$
    \Require $0 < \varepsilon < 1$
    \Ensure 3D tensor \textbf{output}
    \Ensure ponder cost $\rho$
    \State $\mathbf{x} = \textbf{input}$
    \State $c = 0$ \Comment{Cumulative halting score}
    \State $R = 1$ \Comment{Remainder value}
    \State $\textbf{output} = 0$ \Comment{Output of the block}
    \State $\rho = 0$
    \For{$l = 1 \dots L$}
        \State $\mathbf{x} = F^l (\mathbf{x})$
        \If{$l < L$} $h = H^l (\mathbf{x})$
        \Else $\ h = 1$
        \EndIf
        \State{$c \mathrel{{+}{=}} h$}
        \State{$\rho \mathrel{{+}{=}} 1$}
        \If{$c < 1-\varepsilon$}
            \State{$\textbf{output} \mathrel{{+}{=}} h \cdot \mathbf{x}$}
            \State{$R \mathrel{{-}{=}} h$}
        \Else
            \State{$\textbf{output} \mathrel{{+}{=}} R \cdot \mathbf{x}$}
            \State{$\rho \mathrel{{+}{=}} R$}
            \State \textbf{break}
        \EndIf
    \EndFor
    \State \Return $\textbf{output}, \rho$
\end{algorithmic}
\caption{Adaptive Computation Time for one block of residual units. ACT does not require storing the intermediate residual units outputs.}
\label{algorithm:act}
\end{algorithm}

The output of the block is now defined as the outputs of residual units weighted by the halting distribution.
Since representations of residual units are compatible with each other~\cite{huang2016deep,greff2017highway}, the weighted average also produces a feature representation of the same type.
The values of $\mathbf{x}^{N+1}, \dots, \mathbf{x}^{L}$ have zero weight and therefore their evaluation can be skipped:
\begin{equation}
    \textbf{output} = \sum_{l=1}^L p^l \mathbf{x}^l = \sum_{l=1}^N p^l \mathbf{x}^l.
    \label{eqn:output}
\end{equation}

Ideally, we would like to directly minimize the number of evaluated units $N$.
However, $N$ is a piecewise constant function of the halting scores that cannot be optimized with gradient descent.
Instead, we introduce the ponder cost $\rho$, an almost everywhere differentiable upper bound on the number of evaluated units $N$ (recall that $R \geq 0$):
\begin{equation}
    \rho = N + R.
\end{equation}
When differentiating $\rho$, we ignore the gradient of $N$.
Also, note that $R$ is not a continuous function of the halting scores~\cite{larochelle2016notes}.
The discontinuities happen in the configurations of halting scores where $N$ changes value.
Following~\cite{graves2016adaptive}, we ignore these discontinuities and find that they do not impede training.
Algorithm~\ref{algorithm:act} shows the description of ACT.

The partial derivative of the ponder cost \wrt a halting score $h^l$ is
\begin{equation}
    \frac{\partial \rho}{\partial h^l} = \begin{cases}
            -1 &\text{if } l < N, \\
            0 &\text{if } l \geq N.
            \end{cases}
\end{equation}
Therefore, minimizing the ponder cost increases $h^1, \dots, h^{N-1}$, making the computation stop earlier.
This effect is balanced by the original loss function $\mathcal{L}$ which also depends on the halting scores via the block output, eqn. \eqref{eqn:output}.
Intuitively, the more residual units are used, the better the output, so minimizing $\mathcal{L}$ usually increases the weight $R$ of the last used unit's output $\mathbf{x}^N$, which in turn decreases $h^1, \dots, h^{N-1}$. 

ACT has several important advantages.
First, it adds very few parameters and computation to the base model.
Second, it allows to calculate the output of the block ``on the fly'' without storing all the intermediate residual unit outputs and halting scores in memory.
For example, this would not be possible if the halting distribution were a softmax of halting scores, as done in soft attention~\cite{xu2015show}.
Third, we can recover a block with any constant number of units $l \leq L$ by setting $h^1=\dots=h^{l-1}=0, h^l=1$.
Therefore, ACT is a strict generalization of standard ResNet.

We apply ACT to each block independently and then stack the obtained blocks as in the original ResNet.
The input of the next block becomes the weighted average of the residual units from the previous block, eqn. \eqref{eqn:output}.
A similar connectivity pattern has been explored in~\cite{huang2016densely}.
We add the sum of the ponder costs $\rho_k, k=1 \dots K$ from the $K$ blocks to the original loss function $\mathcal{L}$:
\begin{equation}
    \mathcal{L}' = \mathcal{L} + \tau \sum_{k=1}^K \rho_k.
\end{equation}
The resulting loss function $\mathcal{L}'$ is differentiable and can be optimized using conventional backpropagation.
$\tau \geq 0$ is a regularization coefficient which controls the trade-off between optimizing the original loss function and the ponder cost.

\subsection{Spatially Adaptive Computation Time}

\begin{figure}
    \centering
    \includegraphics[width=\linewidth]{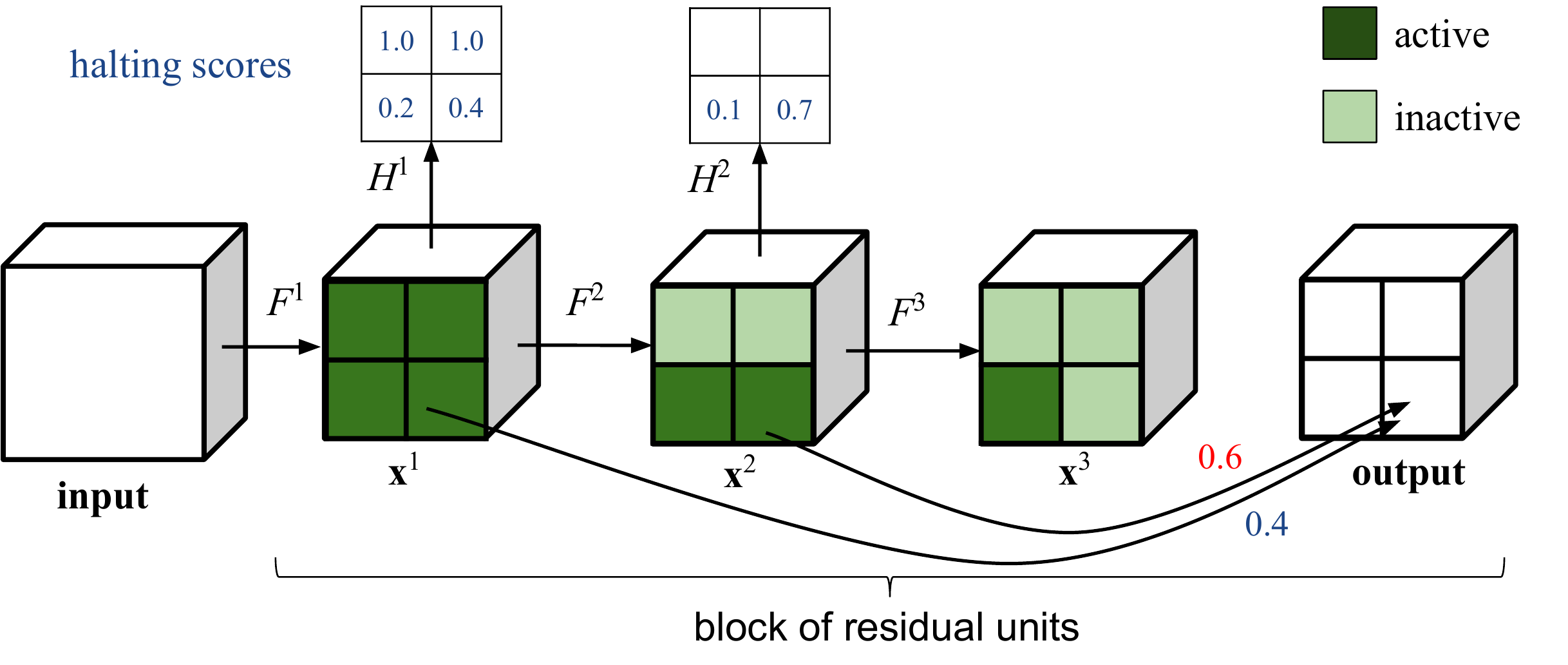}
    \caption{Spatially Adaptive Computation Time (SACT) for one block of residual units. We apply ACT to each spatial position of the block. As soon as position's cumulative halting score reaches 1, we mark it as inactive. See alg.~\ref{algorithm:sact}. SACT learns to choose the appropriate amount of computation for each spatial position in the block.}
    \label{fig:sact}
\end{figure}

In this section, we present Spatially Adaptive Computation Time (SACT).
We adjust the per-position amount of computation by applying ACT to each spatial position of the block, as shown in fig.~\ref{fig:sact}.
As we show in the experiments, SACT can learn to focus the computation on the regions of interest.

\begin{figure}
    \hspace{40pt}
    \includegraphics[width=0.7\linewidth]{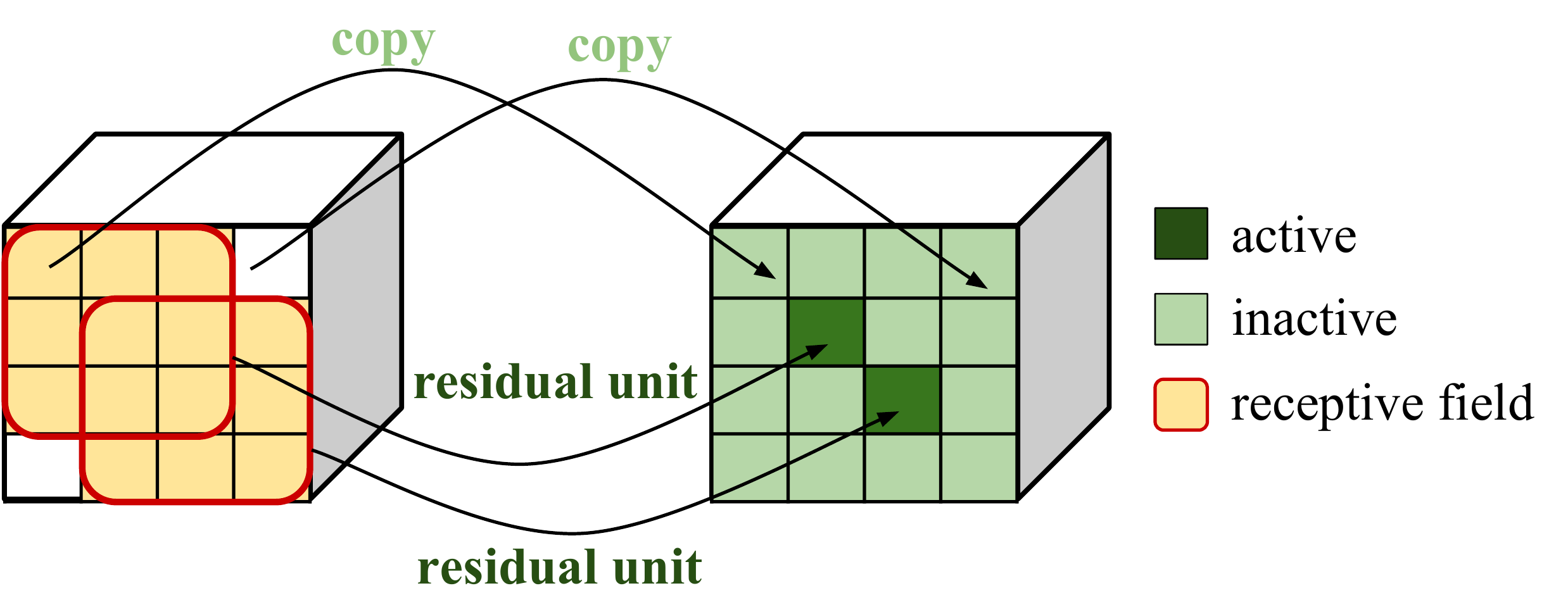}
    \caption{Residual unit with active and inactive positions in SACT. This transformation can be implemented efficiently using the perforated convolutional layer~\cite{figurnov2016perforatedcnns}.}
    \label{fig:sact-active}
\end{figure}

\begin{figure}
    \centering
    \includegraphics[width=0.7\linewidth]{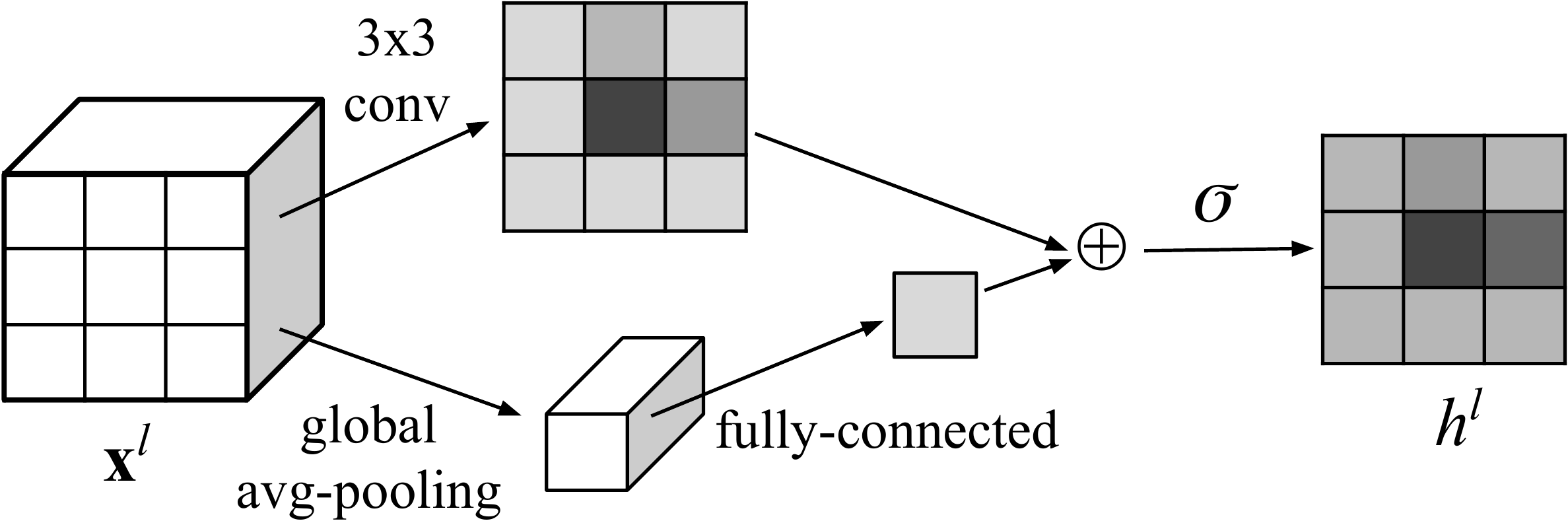}
    \caption{SACT halting scores. Halting scores are evaluated fully convolutionally making SACT applicable to images of arbitrary resolution. SACT becomes ACT if the $3 \times 3$ conv weights are set to zero.}
    \label{fig:sact-halting-score}
\end{figure}

We define the \emph{active positions} as the spatial locations where the cumulative halting score is less than one.
Because an active position might have \emph{inactive} neighbors, the values for the the inactive positions need to be imputed to evaluate the residual unit in the active positions.
We simply copy the previous value for the inactive spatial positions, which is equivalent to setting the residual function $f(\mathbf{x})$ value to zero, as displayed in fig.~\ref{fig:sact-active}.
The evaluation of a block can be stopped completely as soon as all the positions become inactive.
Also, the ponder cost is averaged across the spatial positions to make it comparable with the ACT ponder cost.
The full algorithm is described in alg.~\ref{algorithm:sact}.

\begin{algorithm}
\begin{algorithmic}[1]
    \Require 3D tensor \textbf{input}
    \Require number of residual units in the block $L$
    \Require $0 < \varepsilon < 1$
    \Statex \Comment{input and output have different shapes}
    \Ensure 3D tensor \textbf{output} of shape $H \times W \times C$
    \Ensure ponder cost $\rho$
    \State $\hat{\mathbf{x}} = \textbf{input}$
    \State $\mathcal{X} = \{1 \dots H\} \times \{1 \dots W\}$
    \ForAll{$(i,j) \in \mathcal{X}$}
        \State $a_{ij} = \textbf{true}$ \Comment{Active flag}
        \State $c_{ij} = 0$ \Comment{Cumulative halting score}
        \State $R_{ij} = 1$ \Comment{Remainder value}
        \State $\textbf{output}_{ij} = 0$ \Comment{Output of the block}
        \State $\rho_{ij} = 0$ \Comment{Per-position ponder cost}
	\EndFor
    \For{$l = 1 \dots L$}
        \If{$\textbf{not } a_{ij}\ \forall (i,j) \in \mathcal{X}$}{$\textbf{ break}$} \EndIf
        \ForAll{$(i,j) \in \mathcal{X}$}
            \If{$a_{ij}$} $\mathbf{x}_{ij} = F^l (\hat{\mathbf{x}})_{ij}$
            \Else $\ \mathbf{x}_{ij} = \hat{\mathbf{x}}_{ij}$ \EndIf
        \EndFor
    	\ForAll{$(i,j) \in \mathcal{X}$}
    	    \If{$\textbf{not } a_{ij}$} $\textbf{continue}$ \EndIf
	        \If{$l < L$} $h_{ij} = H^l (\mathbf{x})_{ij}$
	        \Else $\ h_{ij} = 1$
	        \EndIf
	        \State $c_{ij} \mathrel{{+}{=}} h_{ij}$
	        \State $\rho_{ij} \mathrel{{+}{=}} 1$
	        \If{$c_{ij} < 1-\varepsilon$}
	            \State $\textbf{output}_{ij} \mathrel{{+}{=}} h_{ij} \cdot \mathbf{x}_{ij}$
	            \State $R_{ij} \mathrel{{-}{=}} h_{ij}$
	        \Else
	            \State $\textbf{output}_{ij} \mathrel{{+}{=}} R_{ij} \cdot \mathbf{x}_{ij}$
	            \State $\rho_{ij} \mathrel{{+}{=}} R_{ij}$
	            \State $a_{ij} = \textbf{false}$
	        \EndIf
        \EndFor
        \State $\hat{\mathbf{x}} = \mathbf{x}$
    \EndFor
    \State $\rho = \sfrac{\sum_{(i,j) \in \mathcal{X}}\rho_{ij}}{(HW)}$
    \State \Return $\textbf{output},\ \rho$
\end{algorithmic}
\caption{Spatially Adaptive Computation Time for one block of residual units}
\label{algorithm:sact}
\end{algorithm}

We define the halting scores for SACT as
\begin{equation}
    H^l(\mathbf{x}) = \sigma (\widetilde{W}^l \ast \mathbf{x} + W^l \operatorname{pool}(\mathbf{x}) + b^l),
\end{equation}
where $\ast$ denotes a $3 \times 3$ convolution with a single output channel and $\operatorname{pool}$ is a global average-pooling (see fig.~\ref{fig:sact-halting-score}).
SACT is fully convolutional and can be applied to images of any size.

Note that SACT is a more general model than ACT, and, consequently, than standard ResNet.
If we choose $\widetilde{W}^l = 0$, then the halting scores for all spatial positions coincide.
In this case the computation for all the positions halts simultaneously and we recover the ACT model.

SACT requires evaluation of the residual function $f(\mathbf{x})$ in just the active spatial positions.
This can be performed efficiently using the \emph{perforated convolutional layer} proposed in~\cite{figurnov2016perforatedcnns} (with skipped values replaced by zeros instead of the nearest neighbor's values).
Recall that the residual function consists of a stack of $1 \times 1$, $3 \times 3$ and $1 \times 1$ convolutional layers.
The first convolutional layer has to be evaluated in the positions obtained by \textit{dilating} the active positions set with a $3 \times 3$ kernel.
The second and third layers need to be evaluated just in the active positions.

An alternative approach to using the perforated convolutional layer is to \textit{tile} the halting scores map. 
Suppose that we share the values of the halting scores $h^l$ within $k \times k$ tiles.
For example, we can perform pooling of $h^l$ with a kernel size $k \times k$ and stride $k$ and then upscale the results by a factor of $k$.
Then, all positions in a tile have the same active flag, and we can apply the residual unit densely to just the active tiles, reusing the commonly available convolution routines.
$k$ should be sufficiently high to mitigate the overhead of the additional kernel calls and the overlapping computations of the first $1 \times 1$ convolution.
Therefore, tiling is advisable when the SACT is applied to high-resolution images.

\section{Related work}

The majority of the work on increasing the computational efficiency of deep convolutional networks focuses on \textit{static} techniques.
These include decompositions of convolutional kernels~\cite{jaderberg2014lowrank} and pruning of connections~\cite{han2016compression}.
Many of these techniques made their way into the design of the standard deep architectures.
For example, Inception~\cite{szegedy2015inception} and ResNet~\cite{he2016deep,he2016identity} use factorized convolutional kernels.

Recently, several works have considered the problem of varying the amount of computation in computer vision.
Cascaded classifiers~\cite{li2015linet,yang2016exploit} are used in object detection to quickly reject ``easy'' negative proposals.
Dynamic Capacity Networks~\cite{almahairi2016dynamic} use the same amount of computation for all images and use image classification-specific heuristic.
PerforatedCNNs~\cite{figurnov2016perforatedcnns} vary the amount of computation spatially but not between images.
\cite{bengio2016conditional} proposes to tune the amount of computation in a fully-connected network using a REINFORCE-trained policy which makes the optimization problem significantly more challenging.

BranchyNet~\cite{teerapittayanon2016branchynet} is the most similar approach to ours although only applicable to classification problems.
It adds classification branches to the intermediate layers of the network.
As soon as the entropy of the intermediate classifications is below some threshold, the network's evaluation halts.
Our preliminary experiments with a similar procedure based on ACT (using ACT to choose the number of blocks to evaluate) show that it is inferior to using less units per block.

\section{Experiments}

We first apply ACT and SACT models to the image classification task for the ImageNet dataset~\cite{deng2009imagenet}.
We show that SACT achieves a better FLOPs-accuracy trade-off than ACT by directing computation to the regions of interest.
Additionally, SACT improves the accuracy on high-resolution images compared to the ResNet model.
Next, we use the obtained SACT model as a feature extractor in the Faster R-CNN object detection pipeline~\cite{ren2015faster} on the COCO dataset~\cite{lin2014microsoft}.
Again we show that we obtain significantly improved FLOPs-mAP trade-off compared to basic ResNet models.
Finally, we demonstrate that SACT ponder cost maps correlate well with the position of human eye fixations by evaluating them as a visual saliency model on the cat2000 dataset~\cite{bylinksii_cat2000} without any training on this dataset.

\subsection{Image classification (ImageNet dataset)}

First, we train the basic ResNet-50 and ResNet-101 models from scratch using asynchronous SGD with momentum (see the supplementary text for the hyperparameters).
Our models achieve similar performance to the reference implementation\footnote{\href{https://github.com/KaimingHe/deep-residual-networks}{https://github.com/KaimingHe/deep-residual-networks}}.
For a single center $224 \times 224$ resolution crop, the reference ResNet-101 model achieves 76.4\% accuracy, 92.9\% recall@5, while our implementation achieves 76\% and 93.1\%, respectively.
Note that our model is the newer pre-activation ResNet~\cite{he2016identity} and the reference implementation is the post-activation ResNet~\cite{he2016deep}.

We use ResNet-101 as the basic architecture for ACT and SACT models.
Thanks to the end-to-end differentiability and deterministic behaviour, we find the same optimization hyperparameters are applicable for training of ACT and SACT as for the ResNet models.
However, special care needs to be taken to address the \textit{dead residual unit} problem in ACT and SACT models.
Since ACT and SACT are deterministic, the last units in the blocks do not get enough training signal and their parameters become obsolete.
As a result, the ponder cost saved by not using these units overwhelms the possible initial gains in the original loss function and the units are never used.
We observe that while the dead residual units can be recovered during training, this process is very slow.
Note that ACT-RNN~\cite{graves2016adaptive} is not affected by this problem since the parameters for all timesteps are shared.

We find two techniques helpful for alleviating the dead residual unit problem.
First, we initialize the bias of the halting scores units to a negative value to force the model to use the last units during the initial stages of learning.
We use $b^l = -3$ in the experiments which corresponds to initially using $1/\sigma(b^l) \approx 21$ units.
Second, we use a \textit{two-stage training} procedure by initializing the ACT/SACT network's weights from the pretrained ResNet-101 model.
The halting score weights are still initialized randomly.
This greatly simplifies learning of a reasonable halting policy in the beginning of training.

As a baseline for ACT and SACT, we consider a non-adaptive ResNet model with a similar number of floating point operations.
We take the average numbers of units used in each block in the ACT or SACT model (for SACT we also average over the spatial dimensions) and round them to the nearest integers.
Then, we train a ResNet model with such number of units per block.
We follow the two-stage training procedure by initializing the network's parameters with the the first residual units of the full ResNet-101 in each block.
This slightly improves the performance compared to using the random initialization.

\begin{figure}
    \centering
    \begin{subfigure}{0.5\linewidth}
        \centering
        \includegraphics[width=\linewidth]{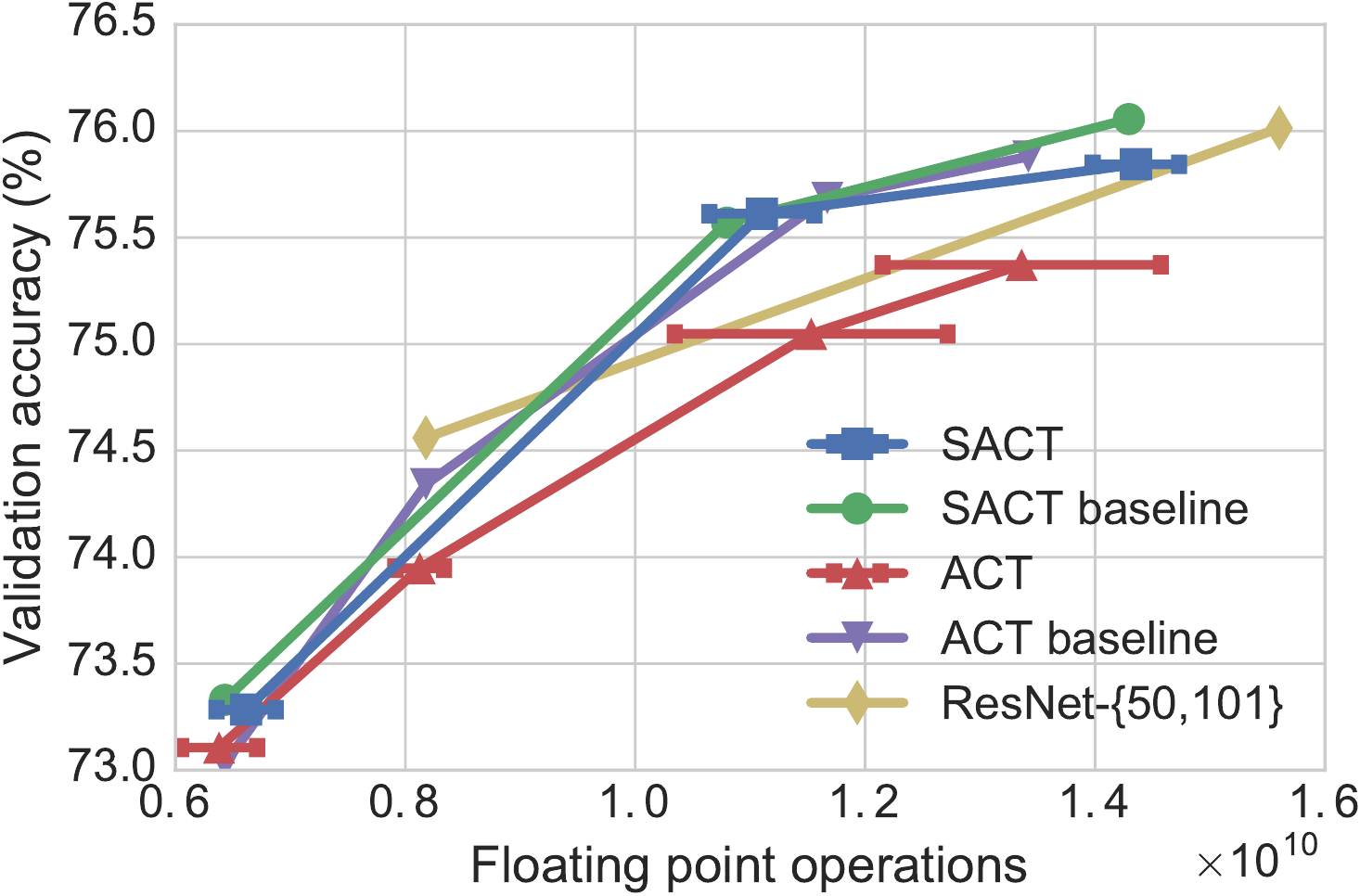}
        \caption{Test resolution $224\times224$}
    \end{subfigure}%
    \begin{subfigure}{0.5\linewidth}
        \centering
        \includegraphics[width=\linewidth]{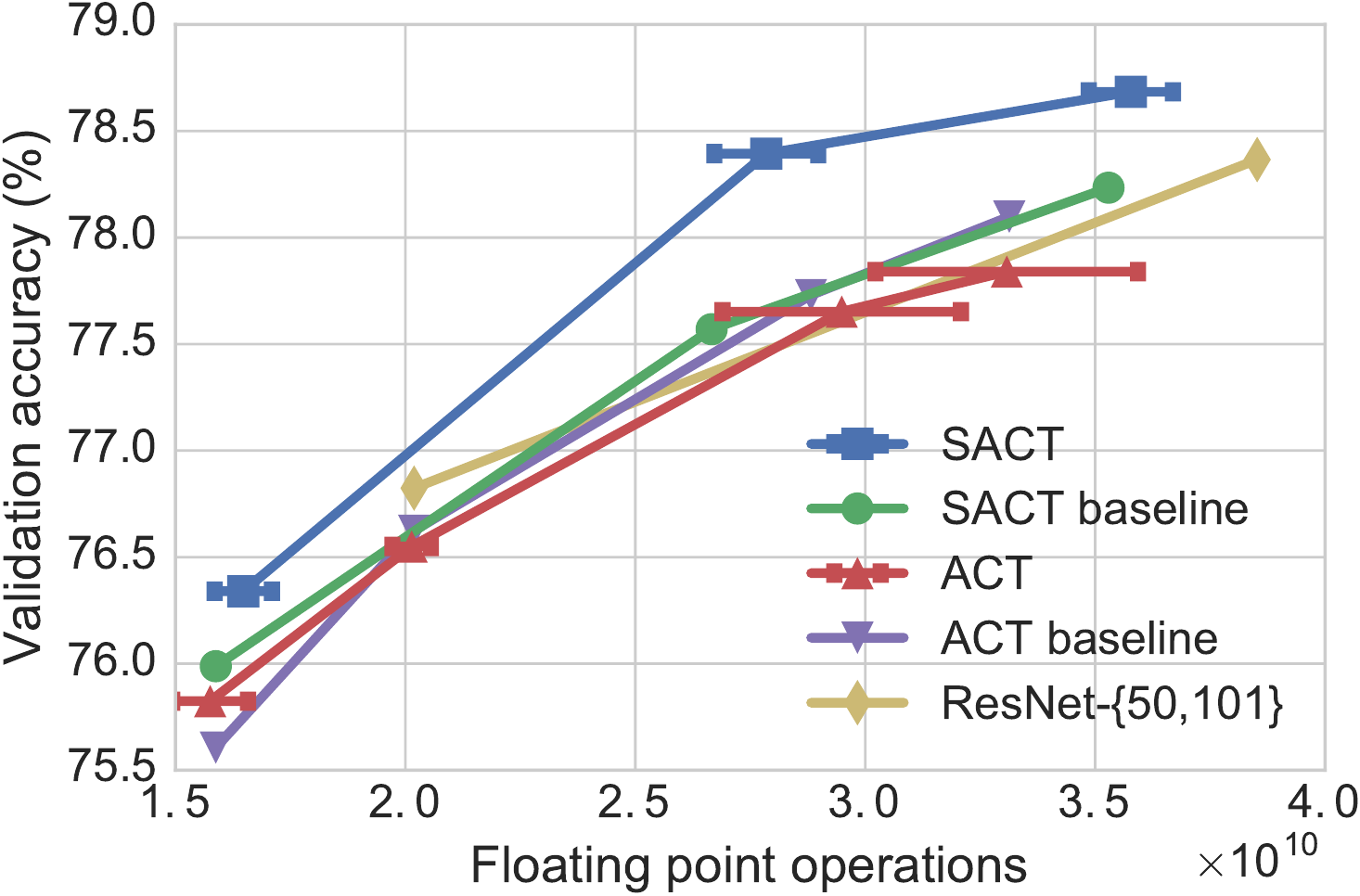}
        \caption{Test resolution $352\times352$}
    \end{subfigure}
    \begin{subfigure}{0.5\linewidth}
        \centering
        \includegraphics[width=\linewidth]{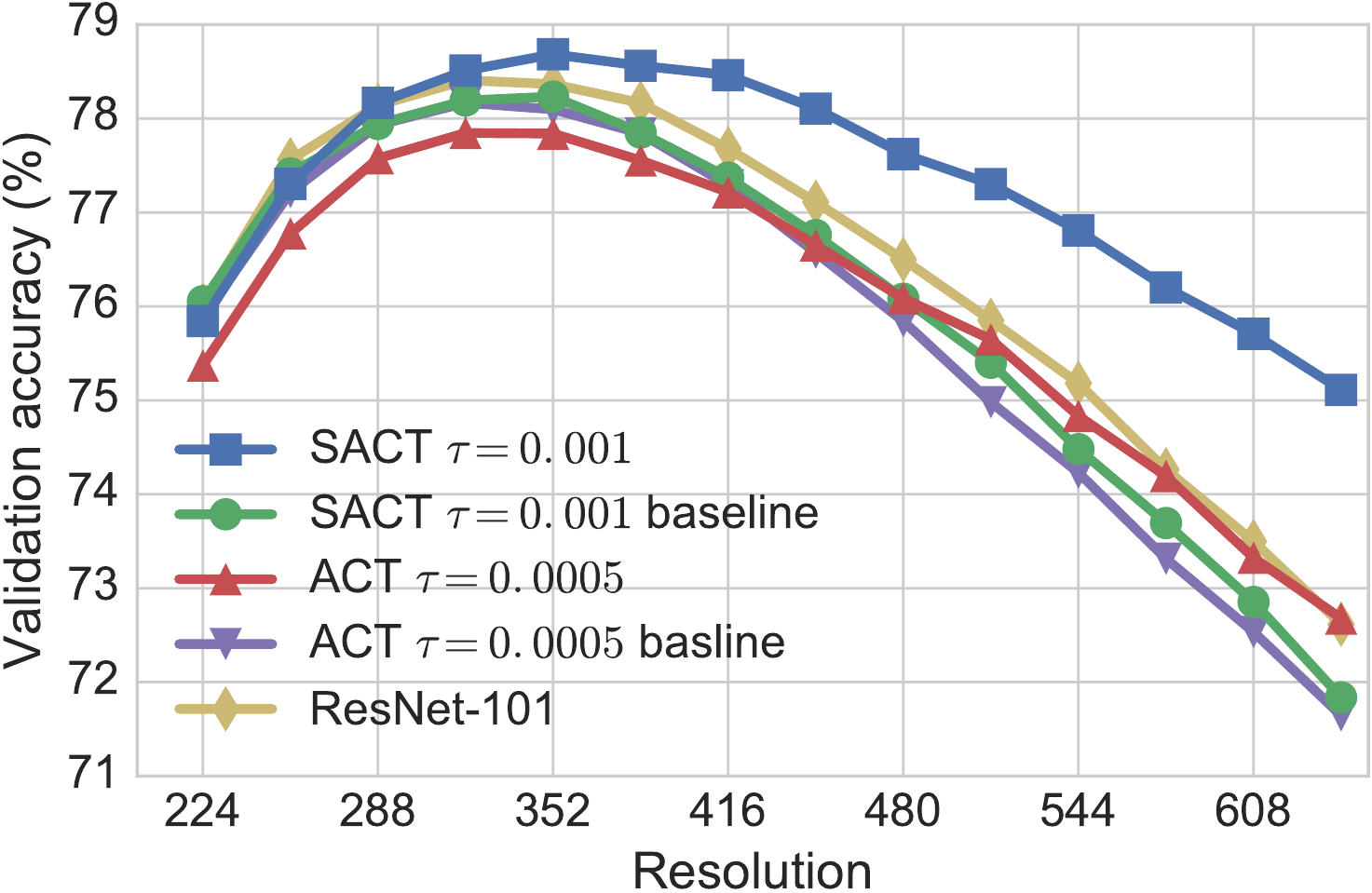}
        \caption{Resolution \vs accuracy}
    \end{subfigure}%
    \begin{subfigure}{0.5\linewidth}
        \centering
        \includegraphics[width=\linewidth]{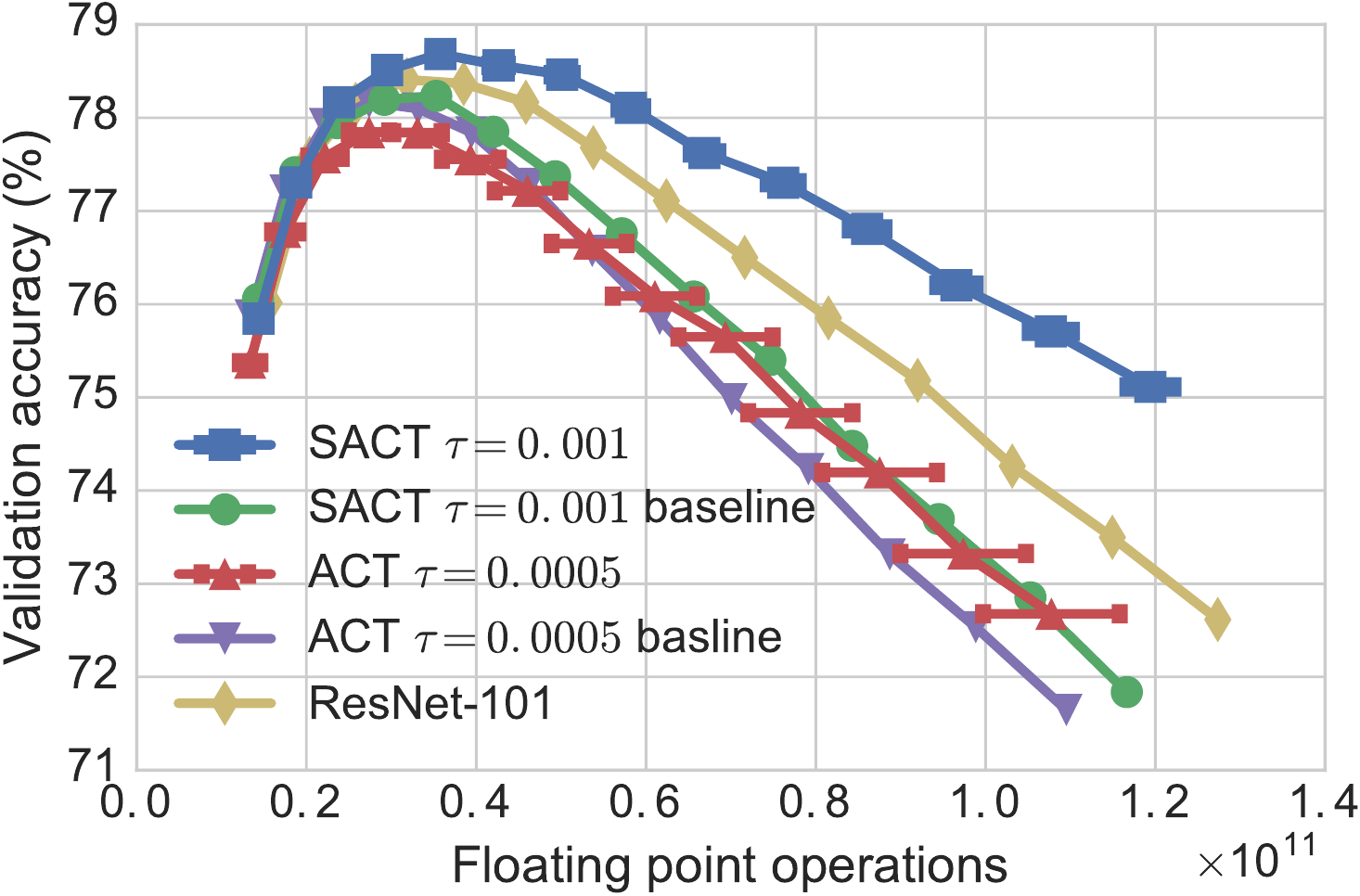}
        \caption{FLOPs \vs accuracy for varying resolution}
    \end{subfigure}
    \caption{ImageNet validation set. Comparison of ResNet, ACT, SACT and the respective baselines. Error bars denote one standard deviation across images. All models are trained with $224 \times 224$ resolution images. SACT outperforms ACT and baselines when applied to images whose resolutions are higher than the training images. The advantage margin grows as resolution difference increases.}
    \label{fig:imagenet-time-accuracy}
\end{figure}

\begin{figure}
    \centering
    \includegraphics[height=1.25cm]{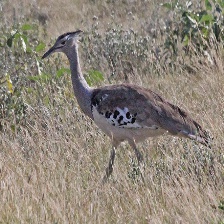}
    \includegraphics[height=1.25cm]{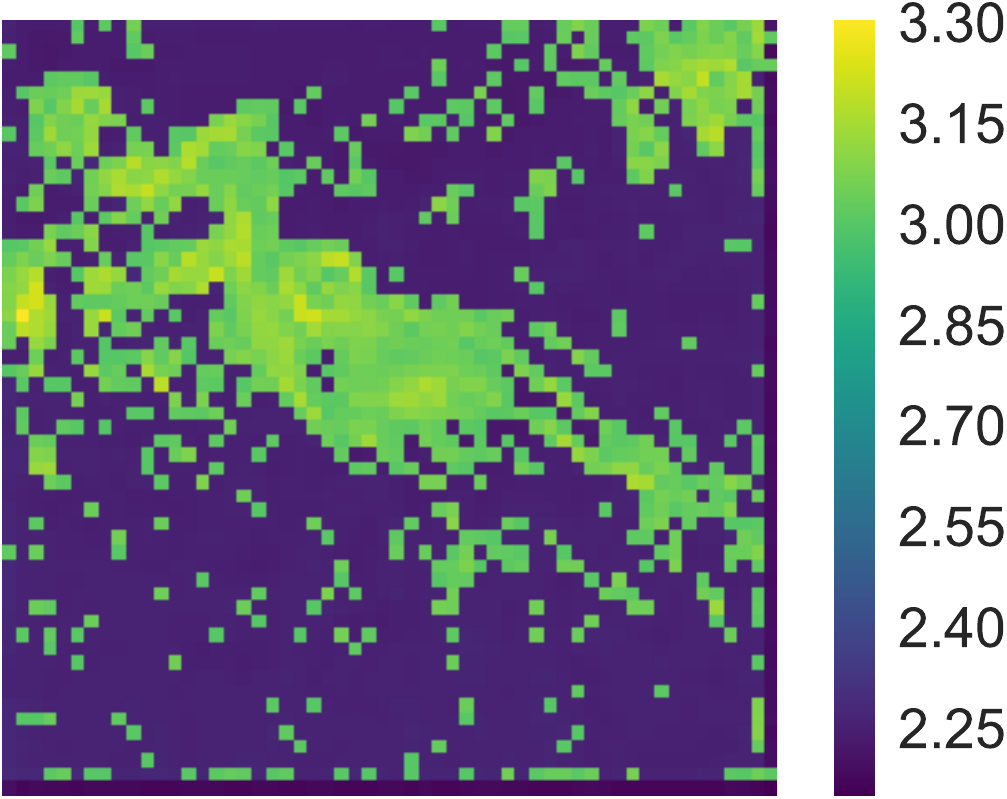}
    \includegraphics[height=1.25cm]{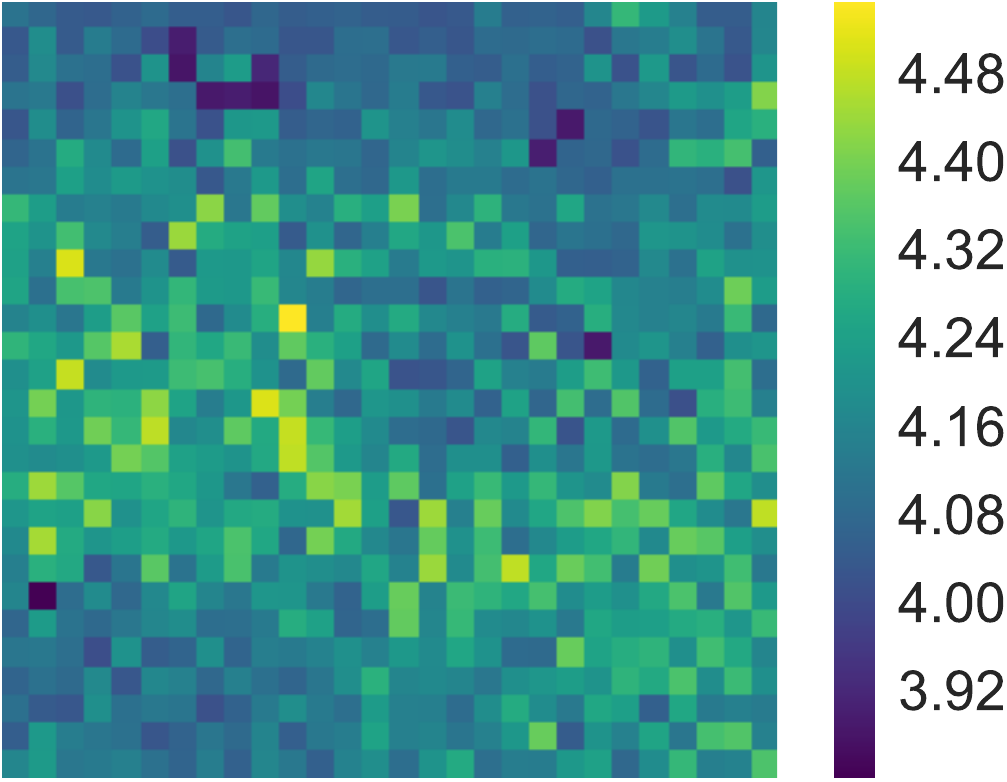}
    \includegraphics[height=1.25cm]{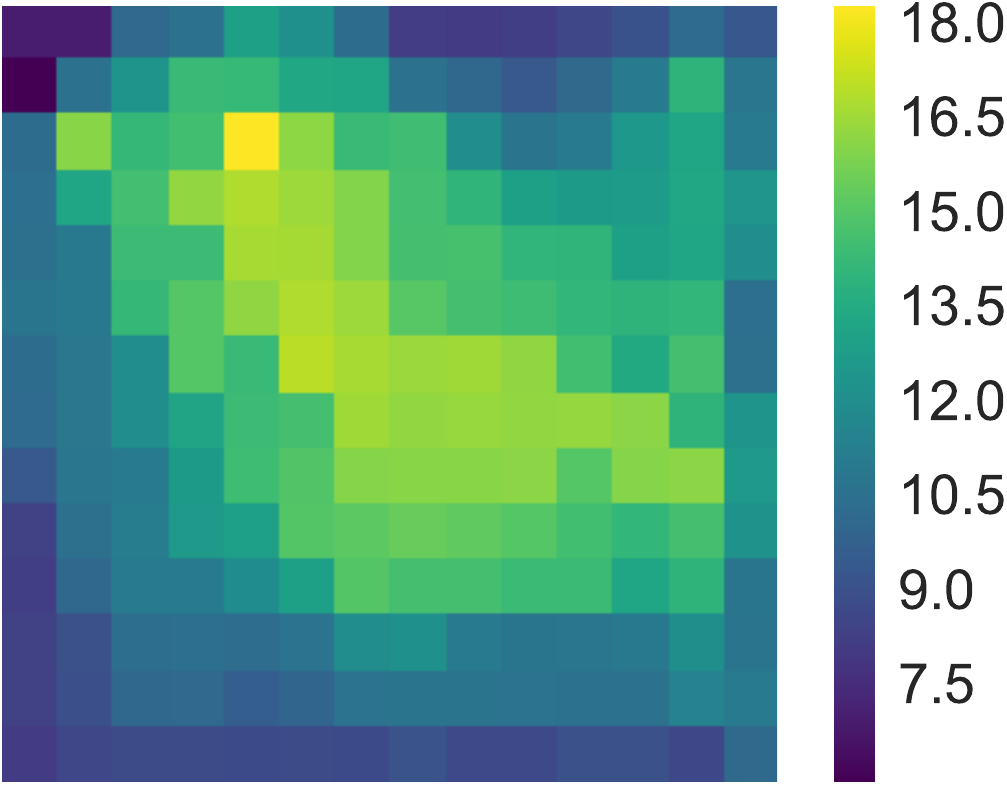}
    \includegraphics[height=1.25cm]{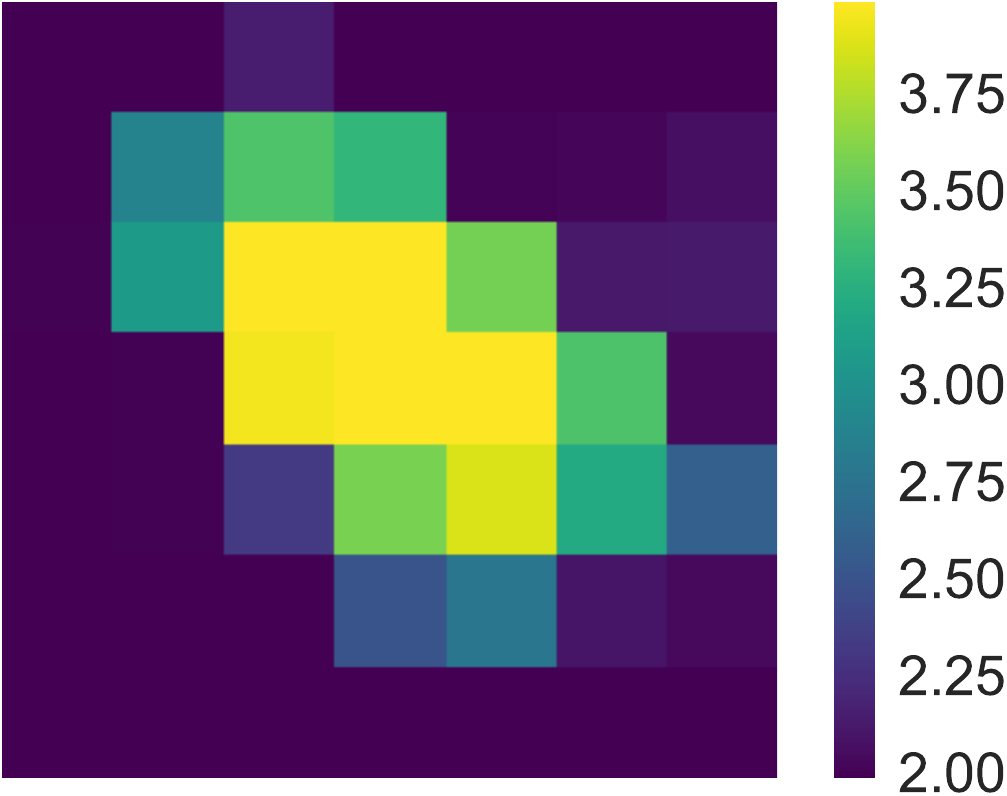}
    \caption{Ponder cost maps for each block (SACT $\tau=0.005$, ImageNet validation image). Note that the first block reacts to the low-level features while the last two blocks attempt to localize the object.}
    \label{fig:imagenet-sact-blocks}
\end{figure}

\begin{figure*}
    \centering
    \begin{subfigure}{0.9\linewidth}
        \centering
        \includegraphics[width=0.115\linewidth]{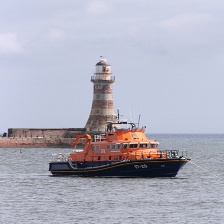}
        \includegraphics[width=0.115\linewidth]{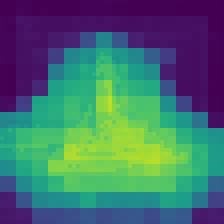}
        \includegraphics[width=0.115\linewidth]{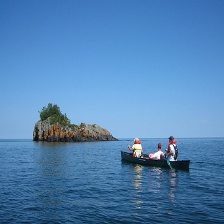}
        \includegraphics[width=0.115\linewidth]{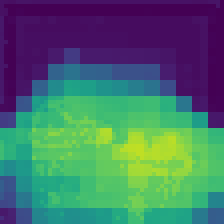}
        \includegraphics[width=0.115\linewidth]{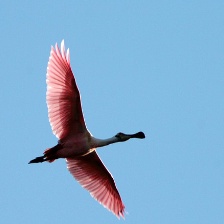}
        \includegraphics[width=0.115\linewidth]{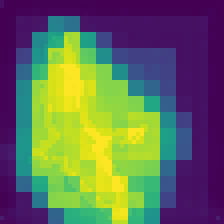}
        \includegraphics[width=0.115\linewidth]{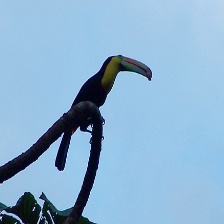}
        \includegraphics[width=0.115\linewidth]{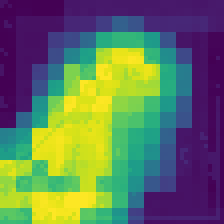}
        \includegraphics[width=0.028\linewidth]{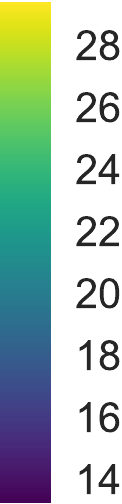}
    \end{subfigure}
    \begin{subfigure}{0.9\linewidth}
        \centering
        \includegraphics[width=0.115\linewidth]{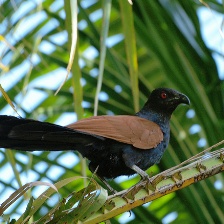}
        \includegraphics[width=0.115\linewidth]{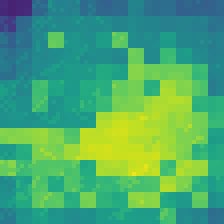}
        \includegraphics[width=0.115\linewidth]{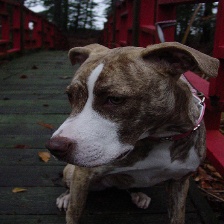}
        \includegraphics[width=0.115\linewidth]{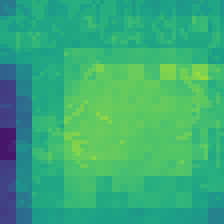}
        \includegraphics[width=0.115\linewidth]{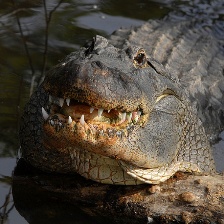}
        \includegraphics[width=0.115\linewidth]{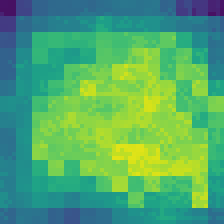}
        \includegraphics[width=0.115\linewidth]{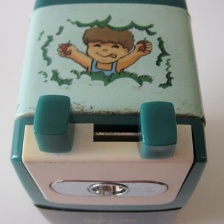}
        \includegraphics[width=0.115\linewidth]{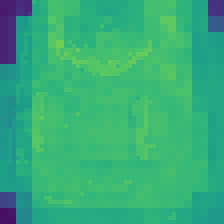}
        \includegraphics[width=0.028\linewidth]{{inet_sact_0.005_colorbar-crop}.pdf}
    \end{subfigure}
    \begin{subfigure}{0.9\linewidth}
        \centering
        \includegraphics[width=0.115\linewidth]{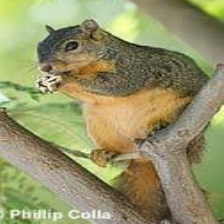}
        \includegraphics[width=0.115\linewidth]{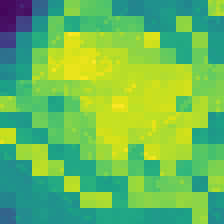}
        \includegraphics[width=0.115\linewidth]{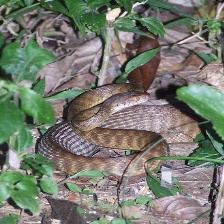}
        \includegraphics[width=0.115\linewidth]{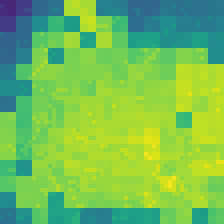}
        \includegraphics[width=0.115\linewidth]{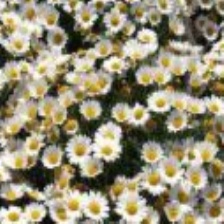}
        \includegraphics[width=0.115\linewidth]{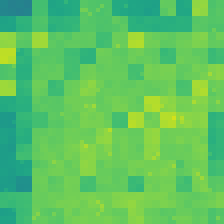}
        \includegraphics[width=0.115\linewidth]{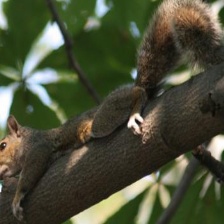}
        \includegraphics[width=0.115\linewidth]{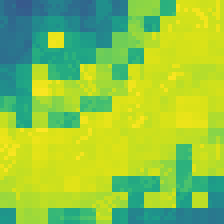}
        \includegraphics[width=0.028\linewidth]{{inet_sact_0.005_colorbar-crop}.pdf}
    \end{subfigure}
    \caption{ImageNet validation set. SACT ($\tau=0.005$) ponder cost maps. Top: low ponder cost (19.8-20.55), middle: average ponder cost (23.4-23.6), bottom: high ponder cost (24.9-26.0). SACT typically focuses the computation on the region of interest.}
    \label{fig:imagenet-sact-visualization}
\end{figure*}

We compare ACT and SACT to ResNet-50, ResNet-101 and the baselines in fig.~\ref{fig:imagenet-time-accuracy}.
We measure the average per-image number of floating point operations (FLOPs) required for evaluation of the validation set.
We treat multiply-add as two floating point operations.
The FLOPs are calculated just for the convolution operations (perforated convolution for SACT) since all other operations (non-linearities, pooling and output averaging in ACT/SACT) have minimal impact on this metric.
The ACT models use $\tau \in \{0.0005, 0.001, 0.005, 0.01\}$ and SACT models use $\tau \in \{0.001, 0.005, 0.01\}$.
If we increase the image resolution at the test time, as suggested in \cite{he2016identity}, we observe that SACT outperforms ACT and the baselines.
Surprisingly, in this setting SACT has higher accuracy than the ResNet-101 model while being computationally cheaper.
Such accuracy improvement does not happen for the baseline models or ACT models.
We attribute this to the improved scale tolerance provided by the SACT mechanism.
The extended results of fig.~\ref{fig:imagenet-time-accuracy}(a,b),
including the average number of residual units per block, are presented in the supplementary.

We visualize the ponder cost for each block of SACT as heat maps (which we call \textit{ponder cost maps} henceforth) in fig.~\ref{fig:imagenet-sact-blocks}.
More examples of the total SACT ponder cost maps are shown in fig.~\ref{fig:imagenet-sact-visualization}.

\subsection{Object detection (COCO dataset)}

Motivated by the success of SACT in classification of high-resolution images and ignoring uninformative background, we now turn to a harder problem of object detection.
Object detection is typically performed for high-resolution images (such as $1000 \times 600$, compared to $224 \times 224$ for ImageNet classification) to allow detection of small objects.
Computational redundancy becomes a big issue in this setting since a large image area is often occupied by the background.

We use the Faster R-CNN object detection pipeline~\cite{ren2015faster} which consists of three stages.
First, the image is processed with a feature extractor. This is the most computationally expensive part.
Second, a Region Proposal Network predicts a number of class-agnostic rectangular proposals (typically 300).
Third, each proposal box's features are cropped from the feature map and passed through a box classifier which predicts whether the proposal corresponds to an object, the class of this object and refines the boundaries.
We train the model end-to-end using asynchronous SGD with momentum, employing Tensorflow's \verb|crop_and_resize| operation, which is 
similar to the Spatial Transformer Network~\cite{jaderberg2015spatial}, to perform cropping of the region proposals.
The training hyperparameters are provided in the supplementary.

We use ResNet blocks 1-3 as a feature extractor and block 4 as a box classifier, as suggested in~\cite{he2016deep}. 
We reuse the models pretrained on the ImageNet classification task and fine-tune them for COCO detection.
For SACT, the ponder cost penalty $\tau$ is only applied to the feature extractor (we use the same value as for ImageNet classification).
We use COCO train for training and COCO val for evaluation
(instead of the combined train+val set which is sometimes used
in the literature).
We do not employ multiscale inference, iterative box refinement or global context.

\begin{figure*}
    \centering
    \begin{subfigure}{0.9\linewidth}
        \centering
        \includegraphics[width=0.235\linewidth]{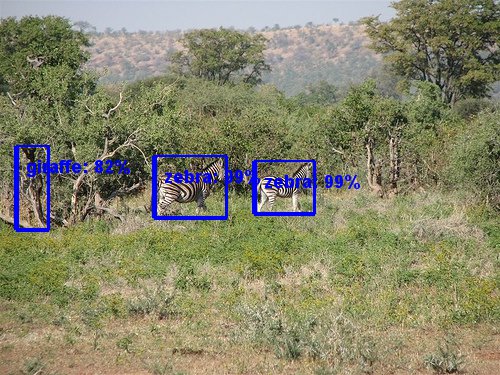}
        \includegraphics[width=0.235\linewidth]{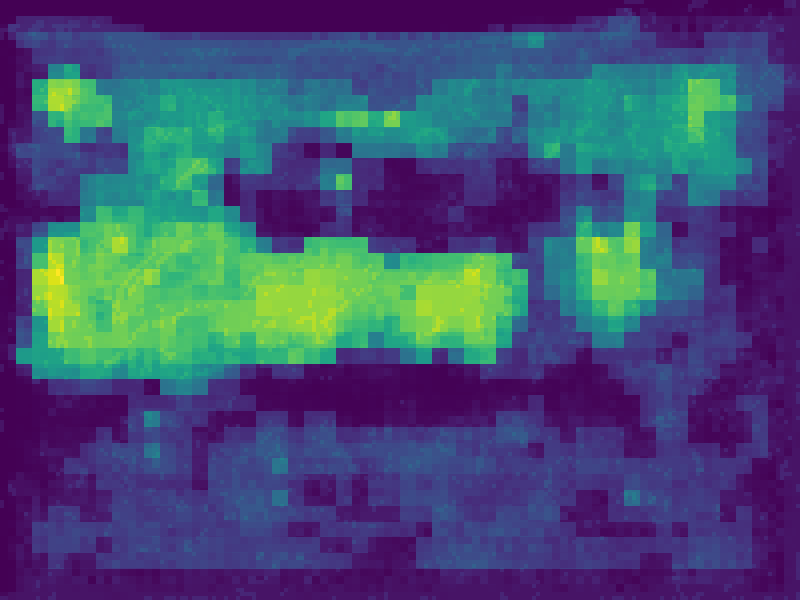}
        \includegraphics[width=0.235\linewidth]{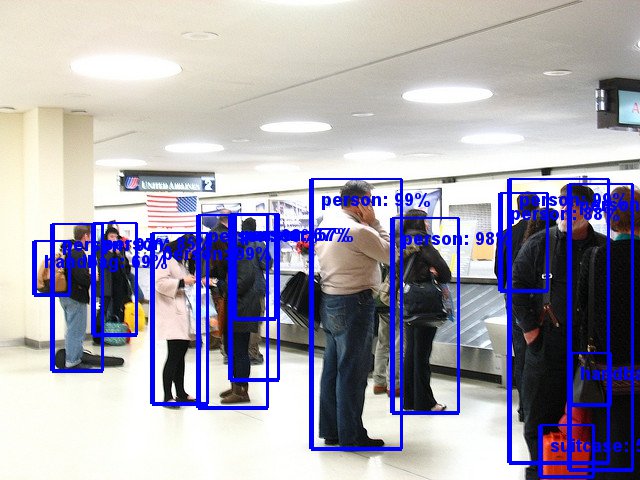}
        \includegraphics[width=0.235\linewidth]{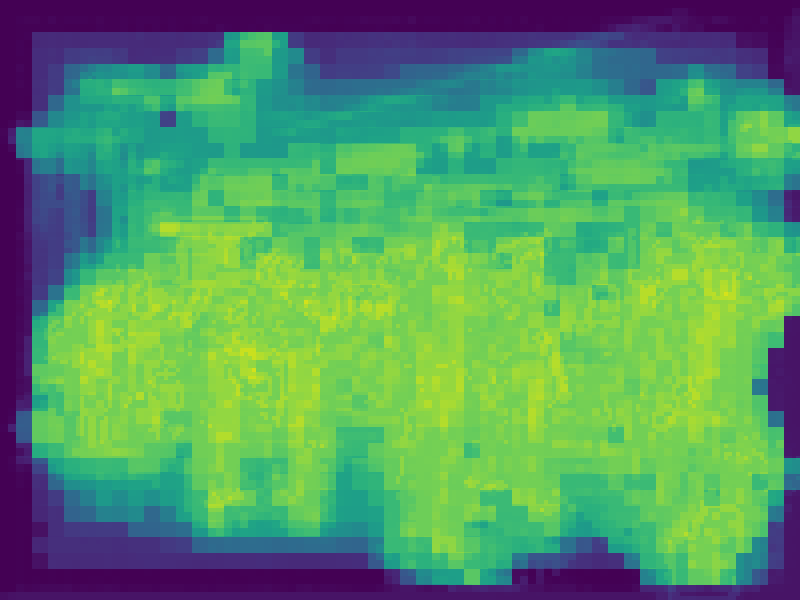}
        \includegraphics[width=0.0275\linewidth]{{coco_sact_0.005_colorbar-crop}.pdf}
    \end{subfigure}
    \begin{subfigure}{0.9\linewidth}
        \centering
        \includegraphics[width=0.235\linewidth]{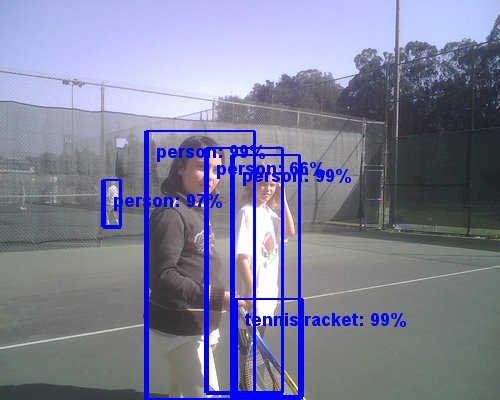}
        \includegraphics[width=0.235\linewidth]{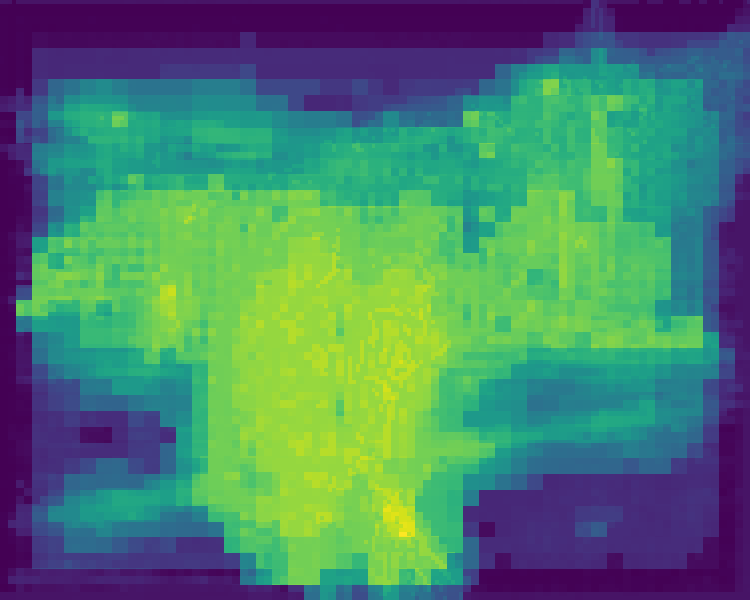}
        \includegraphics[width=0.235\linewidth]{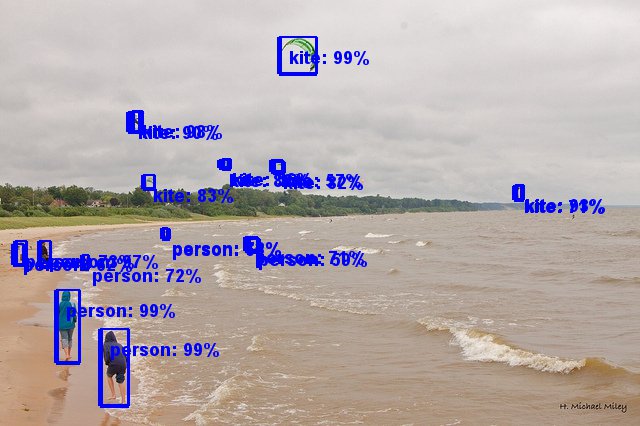}
        \includegraphics[width=0.235\linewidth]{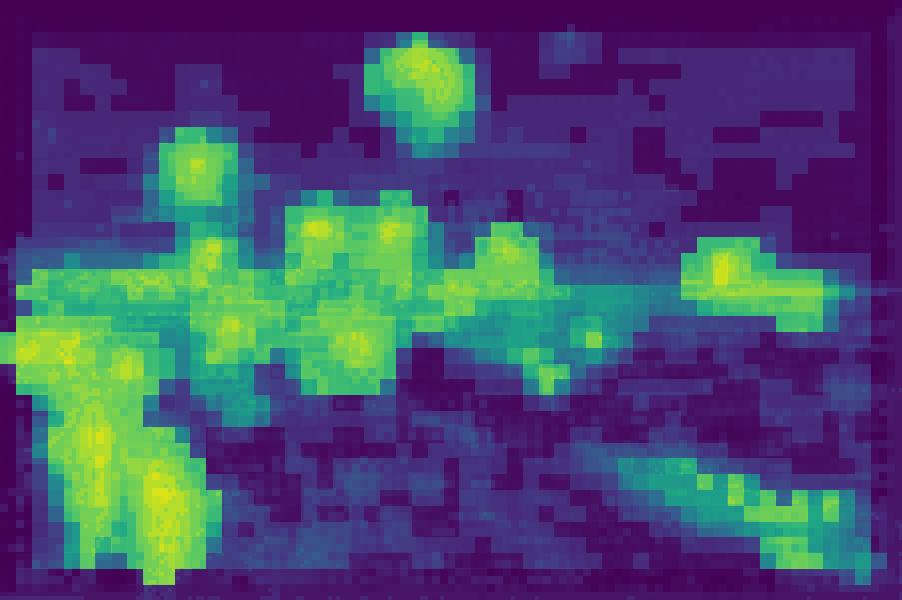}
        \includegraphics[width=0.0245\linewidth]{{coco_sact_0.005_colorbar-crop}.pdf}
    \end{subfigure}
    \caption{COCO testdev set. Detections and feature extractor ponder cost maps ($\tau=0.005$). SACT allocates much more computation to the object-like regions of the image.}
    \label{fig:coco-sact}
\end{figure*}

We find that SACT achieves superior speed-mAP trade-off compared to the baseline of using non-adaptive ResNet as a feature extractor (see table~\ref{table:coco-sact}).
SACT $\tau=0.005$ model has slightly higher FLOPs count than ResNet-50 and $2.1$ points better mAP.
Note that this SACT model outperforms the originally reported result for ResNet-101, $27.2$ mAP~\cite{he2016deep}.
Several examples are presented in fig.~\ref{fig:coco-sact}.

\begin{table}
\begin{center}
\bgroup
\def\arraystretch{1.1}
\tabcolsep=0.10cm
\scalebox{0.9}{
\begin{tabular}{ccc}
\toprule
 Feature extractor & FLOPs (\%) & mAP @ $[.5, .95]$ (\%) \\ \hline
ResNet-101 \cite{he2016deep} & $100$ & $27.2$ \\ \hline
ResNet-50 (our impl.) & $46.6$ & $25.56$ \\
SACT $\tau=0.005$ & $\mathbf{56.0} \pm 8.5$ & $\mathbf{27.61}$ \\
SACT $\tau=0.001$ & $72.4 \pm 8.4$ & $29.04$ \\
ResNet-101 (our impl.) & $100$ & $29.24$ \\
\bottomrule
\end{tabular}
}
\egroup
\end{center}
\caption{COCO val set. Faster R-CNN with SACT results. FLOPs are average ($\pm$ one standard deviation) feature extractor floating point operations relative to ResNet-101 (that does 1.42E+11 operations). SACT improves the FLOPs-mAP trade-off compared to using ResNet without adaptive computation.}
\label{table:coco-sact}
\end{table}

\subsection{Visual saliency (cat2000 dataset)}

We now show that SACT ponder cost maps correlate well with human attention.
To do that, we use a large dataset of visual saliency: the cat2000 dataset~\cite{bylinksii_cat2000}.
The dataset is obtained by showing 4,000 images of 20 scene categories to 24 human subjects and recording their eye fixation positions.
The ground-truth saliency map is a heat map of the eye fixation positions.
We do not train the SACT models on this dataset and simply reuse the ImageNet- and COCO-trained models.
Cat2000 saliency maps exhibit a strong center bias.
Most images contain a blob of saliency in the center even when there is no object of interest located there.
Since our model is fully convolutional, we cannot learn such bias even if we trained on the saliency data.
Therefore, we combine our ponder cost maps with a constant center-biased map.

We resize the $1920\times1080$ cat2000 images to $320\times180$ for ImageNet model and to $640\times360$ for COCO model and pass them through the SACT model.
Following \cite{bylinksii_cat2000}, we consider a linear combination of the Gaussian blurred ponder cost map normalized to $[0, 1]$ range and a ``center baseline,'' a Gaussian centered at the middle of the image.
Full description of the combination scheme is provided in the supplementary.
The first half of the training set images for every scene category is used for determining the optimal values of the Gaussian blur kernel size and the center baseline multiplier, while the second half is used for validation.

\begin{figure}
    \centering
    \begin{subfigure}{0.9\linewidth}
        \centering
        \includegraphics[width=0.32\linewidth,trim={80px 0 80px 0},clip]{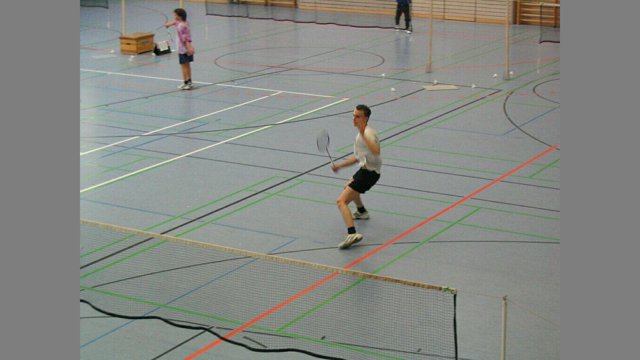}
        \includegraphics[width=0.32\linewidth,trim={80px 0 80px 0},clip]{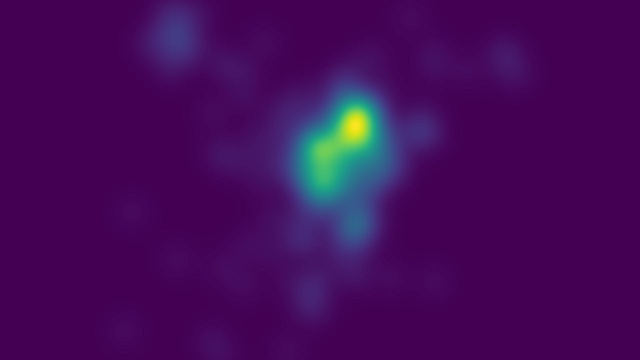}
        \includegraphics[width=0.32\linewidth,trim={80px 0 80px 0},clip]{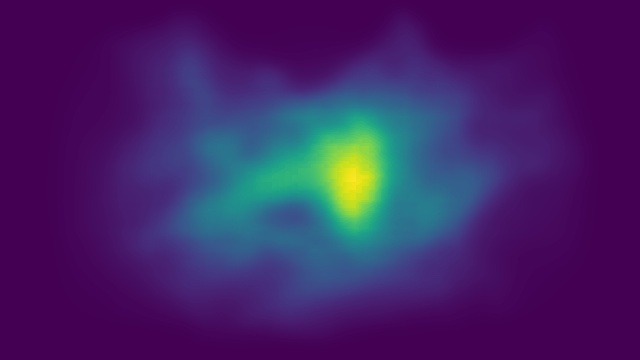}
    \end{subfigure}
    \begin{subfigure}{0.9\linewidth}
        \centering
        \includegraphics[width=0.32\linewidth,trim={80px 0 80px 0},clip]{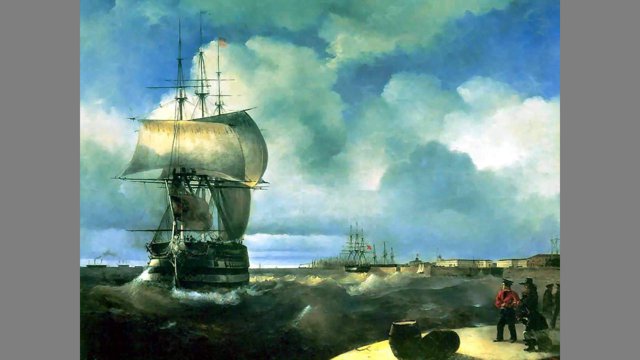}
        \includegraphics[width=0.32\linewidth,trim={80px 0 80px 0},clip]{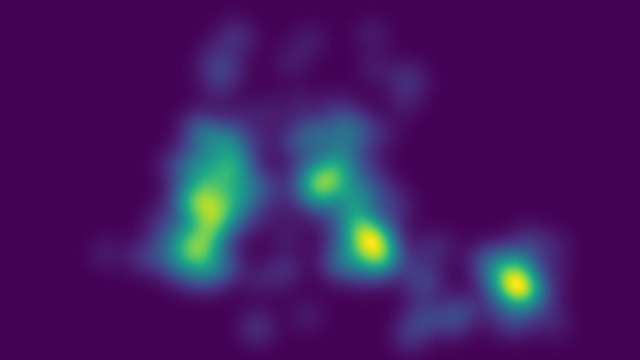}
        \includegraphics[width=0.32\linewidth,trim={80px 0 80px 0},clip]{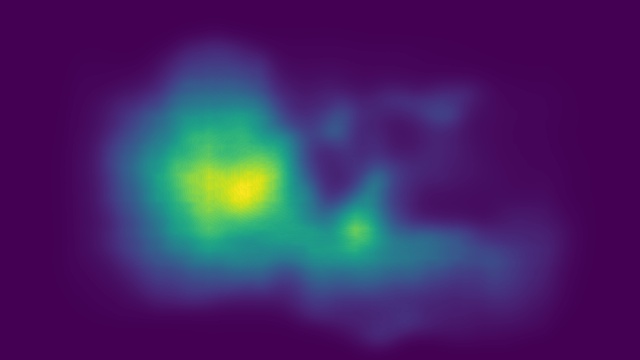}
    \end{subfigure}
    \begin{subfigure}{0.9\linewidth}
        \centering
        \includegraphics[width=0.32\linewidth,trim={80px 0 80px 0},clip]{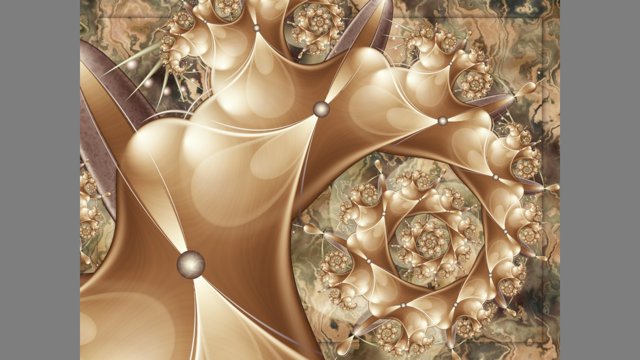}
        \includegraphics[width=0.32\linewidth,trim={80px 0 80px 0},clip]{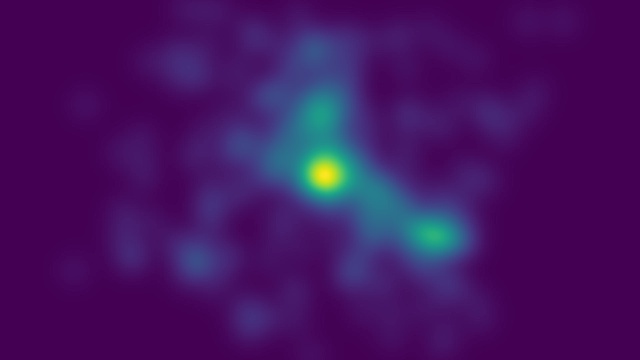}
        \includegraphics[width=0.32\linewidth,trim={80px 0 80px 0},clip]{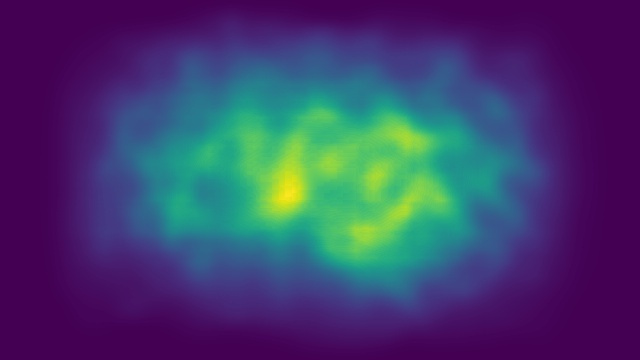}
    \end{subfigure}
    \caption{cat2000 saliency dataset. Left to right: image, human saliency, SACT ponder cost map (COCO model, $\tau=0.005$) with postprocessing (see text) and softmax with temperature $1/5$. Note the center bias of the dataset. SACT model performs surprisingly well on out-of-domain images such as art and fractals.
    }
    \label{fig:cat2000-saliency}
\end{figure}

Table~\ref{table:cat2000-sact} presents the AUC-Judd~\cite{bylinskii2016attentionmetrics} metric, the area under the ROC-curve for the saliency map as a predictor for eye fixation positions.
SACT outperforms the na\"{i}ve center baseline.
Compared to the state-of-the-art deep model DeepFix \cite{kruthiventi2015deepfix} method, SACT does competitively.
Examples are shown in fig.~\ref{fig:cat2000-saliency}.

\begin{table}
\centering
\bgroup
\def\arraystretch{1.1}
\tabcolsep=0.10cm
\scalebox{0.9}{
\begin{tabular}{cc}
\toprule
 Model & AUC-Judd (\%) \\
\hline
Center baseline \cite{bylinksii_cat2000} & $83.4$ \\
DeepFix \cite{kruthiventi2015deepfix} & $87^\dagger$ \\
``Infinite humans'' \cite{bylinksii_cat2000} & $90^\dagger$ \\ \hline
ImageNet SACT $\tau=0.005$ & $84.6$ \\
COCO SACT $\tau=0.005$ & $84.7$ \\
\bottomrule
\vspace{-1em}
\end{tabular}
}
\egroup
\caption{cat2000 validation set. ${}^\dagger$ -  results for the test set. SACT ponder cost maps work as a visual saliency model even without explicit supervision.}
\label{table:cat2000-sact}
\end{table}

\section{Conclusion}

We present a Residual Network based model with a spatially varying computation time.
This model is end-to-end trainable, deterministic and can be viewed as a black-box feature extractor.
We show its effectiveness in image classification and object detection problems.
The amount of per-position computation in this model correlates well with the human eye fixation positions, suggesting that this model captures the important parts of the image.
We hope that this paper will lead to a wider adoption of attention and adaptive computation time in large-scale computer vision systems.
The source code is available at {\small \url{https://github.com/mfigurnov/sact}}.

{\small
\textbf{Acknowledgments.}
D. Vetrov is supported by Russian Academic Excellence Project `5-100'.
R. Salakhutdinov is supported in part by ONR grants N00014-13-1-0721, N00014-14-1-0232, and the ADeLAIDE grant FA8750-16C-0130-001.
}

{\small
\bibliographystyle{ieee}
\bibliography{egbib}
}


\appendix
\onecolumn

\begin{center}
\textbf{\Large Supplementary materials}
\end{center}

In the supplementary materials we present the implementation details and additional experimental results.

\section{Implementation details}
\subsection{Image classification (ImageNet)}
\textbf{Optimization hyperparameters.}
ResNet, ACT and SACT use the same hyperparameters.
We train the networks with 50 workers running asynchronous SGD with momentum $0.9$, weight decay $0.0001$ and batch size $32$.
The training is halted upon convergence after $150-160$ epochs.
Learning rate is initially set to $0.05$ and lowered by a factor of $10$ after every $30$ epochs.
Batch normalization parameters are: epsilon 1e-5, moving average decay $0.997$.
The parameters of the network are initialized with a variance scaling initializer \cite{he2015delving}.

\textbf{Data augmentation.}
We use the Inception v3 data augmentation procedure\footnote{\url{https://github.com/tensorflow/models/blob/master/inception/inception/image_processing.py}} which includes horizontal flipping, scale, aspect ratio, color augmentation.
For the ImageNet images with a provided bounding box, we perform cropping based on the distorted bounding box.
For evaluation, we take a single central crop of $87.5\%$ of the original image's area and then resize this crop to the target resolution.

\subsection{Object detection (COCO)}
ResNet and SACT models use the same hyperparameters.
The images are upscaled (preserving aspect ratio) so that the smaller side is at least 600 pixels.
For data augmentation, we use random horizontal flipping as is described in \cite{ren2015faster}.
We do not employ atrous convolution algorithm.

\textbf{Optimization hyperparameters.}
We use distributed training with 9 workers 
running asynchronous SGD with momentum $0.9$ and a batch size of $1$.
The learning rate is initially set to $0.0003$ and lowered by a factor of 10 after the 800 thousandth and 1 millionth training iterations (batches).
The training proceeds for a total of 1.2 million iterations.
Batch normalization parameters are fixed to the values obtained on ImageNet during training.  

\textbf{Faster R-CNN hyperparameters.}
Other than the training method, our hyperparameters for the Faster 
R-CNN model closely follow those recommended by the original paper \cite{ren2015faster}.
The anchors are generated in the same way, sampled from a regular grid of stride 16.
One change relative to the original paper is the addition of an additional anchor size, so full set of anchor box sizes are
$\{64, 128, 256, 512\}$, with the height and width also varying for each choice of the aspect ratios $\{0.5, 1, 2\}$.
We use 300 object proposals per image.
Non-maximum suppression is performed with $0.6$ IoU threshold.
For each proposal, the features are cropped into a $28\times28$ box with \verb|crop_and_resize| TensorFlow operation, then pooled
to $7\times 7$.

\subsection{Visual saliency (cat2000)}
Here we describe the postprocessing procedure used in the visual saliency experiments.
Consider a ponder cost map $\rho_{ij},\ i \in \{1, \dots, H\},\ j \in \{1, \dots, W\}$.
Let $G_s$ be a Gaussian filter with a standard deviation $s$.

We first normalize this map to $[0, 1]$:
\begin{equation}
    \rho^{n}_{ij} = \frac{\rho_{ij} - \rho_{\min}}{\rho_{\max} - \rho_{\min}},\ i \in \{1, \dots, H\},\ j \in \{1, \dots, W\}.
\end{equation}
where
\begin{equation}
    \rho_{\min} = \min_{ij} \rho_{ij}
\end{equation}
and similarily for $\max$.

Then, we blur the ponder cost map by convolving the Gaussian filter with $\rho$:
\begin{equation}
    \rho^{nb} = G_s \ast \rho^{n}.
\end{equation}

We obtain a baseline map $B_{ij}$ by rescaling the reference centered Gaussian\footnote{\url{https://github.com/cvzoya/saliency/blob/master/code_forOptimization/center.mat}} to $H \times W$ resolution.
We use this map with a weight $\gamma > 0$.

Finally, the postprocessed ponder cost map is defined as a normalized blurred ponder cost map plus the weighted center baseline map:
\begin{equation}
    \rho^{nbc}_{ij} = \rho^{nb}_{ij} + \gamma B_{ij},\ i \in \{1, \dots, H\},\ j \in \{1, \dots, W\}.
\end{equation}

$\rho^{nbc}$ depends on two hyperparameters: the Gaussian filter standard deviation $s$ and baseline map weight $\gamma$.
The values are tuned via grid search.
In the experiments in the paper, we use $s=10$, $\gamma = 0.005$ for both models.

\section{Additional ImageNet results}

We present the extended results of ACT, SACT and ResNet models on ImageNet validation set, including the number of residual units per block, in table \ref{table:imagenet-results}.

\begin{figure}
    \centering
    \begin{subfigure}{0.35\linewidth}
        \centering
        \includegraphics[width=\linewidth]{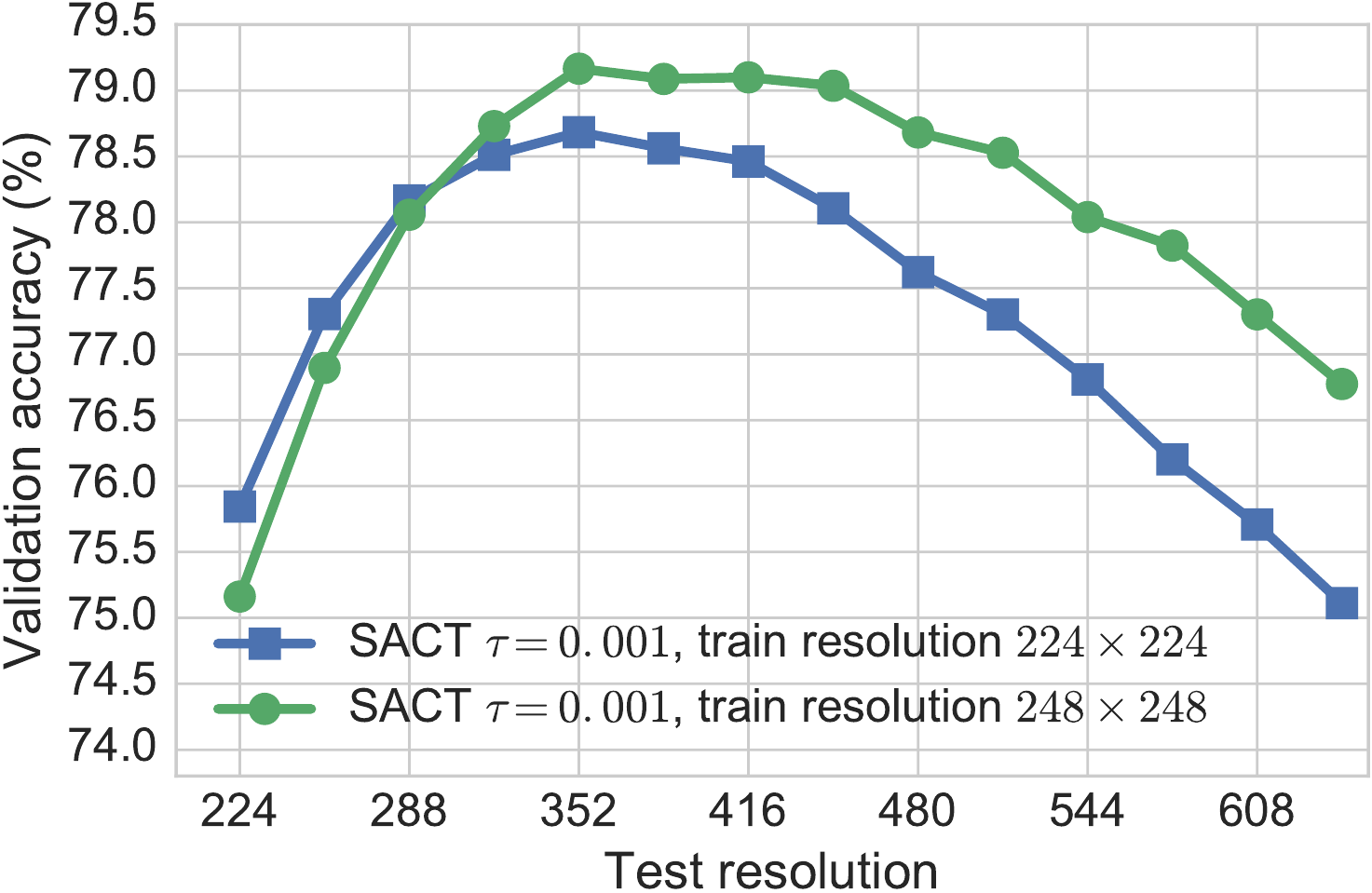}
        \caption{Resolution \vs accuracy}
    \end{subfigure}
    \begin{subfigure}{0.35\linewidth}
        \centering
        \includegraphics[width=\linewidth]{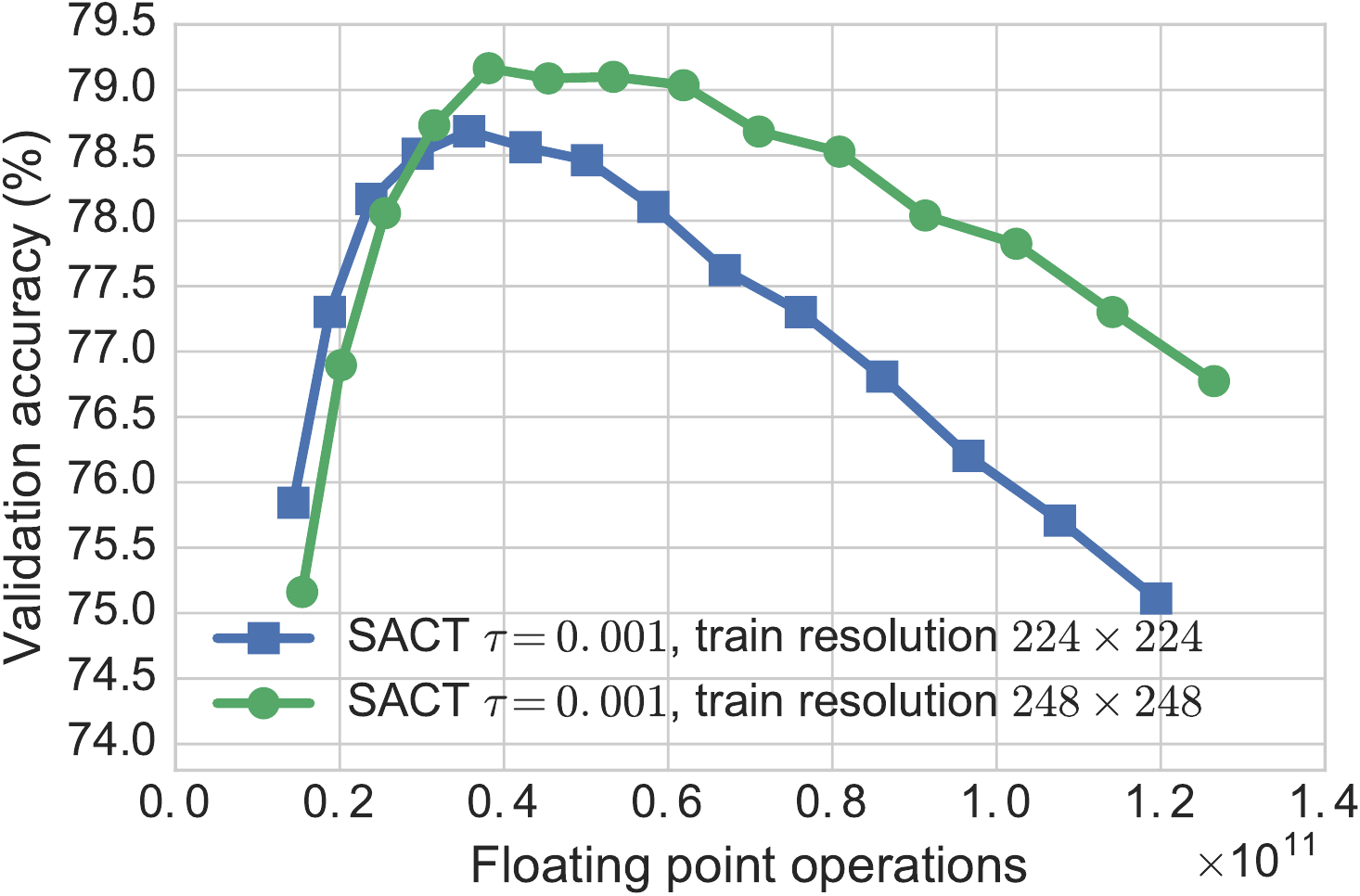}
        \caption{FLOPs \vs accuracy for varying test resolution}
    \end{subfigure}
    \caption{ImageNet validation set. Comparison of SACT with different training resolutions.}
    \label{fig:imagenet-248}
\end{figure}

The ImageNet models in the paper are trained with $224\times224$ images.
Even though all the models are fully convolutional and can be applied to images of any resolution during test time, increasing the training resolution can improve the quality of the model at the cost of longer training and higher GPU memory requirements.
We have explored training of SACT model with a resolution of $248\times248$.
This resolution is the highest we could fit into GPU memory for a batch size of 32 (decreasing the batch size deteriorates the accuracy).
Comparison of SACT models trained with resolutions $224\times224$ and $248\times248$ is presented on fig. \ref{fig:imagenet-248}.
Interestingly, both considered models achieve the highest accuracy at test resolution $352\times352$.
This accuracy is $78.68\%$ for the training resolution of $224\times224$ and $79.16\%$ for the training resolution of $248\times248$.
In the second case, the FLOPs are $6.6\%$ higher.
When the training resolution is increased, the accuracy at lower resolutions is predictably worse, while the accuracy at higher resolutions is better.
The results suggests that training at higher resolutions is beneficial, and that the improved scale tolerance is not diminished when the training resolution is increased.

\begin{table}
\bgroup
\def\arraystretch{1.1}
\tabcolsep=0.10cm
\begin{subtable}{\linewidth}
\centering
\scalebox{0.9}{
\begin{tabular}{lcccc}
Network & FLOPs & Residual units & Accuracy & Recall@5 \\
\hline
ResNet-50 & $\num{8.18E+09}$ & $3, 4, 6, 3$ & $74.56\%$ & $92.37\%$ \\
ResNet-101 & $\num{1.56E+10}$ & $3, 4, 23, 3$ & $76.01\%$ & $93.15\%$ \\
\hline
ACT $\tau = 0.01$ & $\num{6.38E+09} \pm \num{3.31E+08}$ & $2.9 \pm 0.3, 2.7 \pm 0.5, 3.3 \pm 0.4, 3.0 \pm 0.0$ & $73.11\%$ & $91.52\%$ \\
Baseline & $\num{6.43E+09}$ & $3, 3, 3, 3$ & $73.03\%$ & $91.68\%$ \\
\hline
ACT $\tau = 0.005$ & $\num{8.12E+09} \pm \num{2.12E+08}$ & $3.0 \pm 0.0, 4.0 \pm 0.1, 5.9 \pm 0.5, 3.0 \pm 0.0$ & $73.95\%$ & $92.01\%$ \\
Baseline & $\num{8.18E+09}$ & $3, 4, 6, 3$ & $74.34\%$ & $92.19\%$ \\
\hline
ACT $\tau = 0.001$ & $\num{1.15E+10} \pm \num{1.19E+09}$ & $3.0 \pm 0.0, 4.0 \pm 0.0, 13.7 \pm 2.7, 3.0 \pm 0.0$ & $75.05\%$ & $92.58\%$ \\
Baseline & $\num{1.17E+10}$ & $3, 4, 14, 3$ & $75.69\%$ & $93.02\%$ \\
\hline
ACT $\tau = 0.0005$ & $\num{1.34E+10} \pm \num{1.21E+09}$ & $3.0 \pm 0.0, 4.0 \pm 0.0, 17.9 \pm 2.8, 3.0 \pm 0.0$ & $75.37\%$ & $92.76\%$ \\
Baseline & $\num{1.34E+10}$ & $3, 4, 18, 3$ & $75.88\%$ & $93.02\%$ \\
\hline
SACT $\tau = 0.01$ & $\num{6.61E+09} \pm \num{2.57E+08}$ & $2.6 \pm 0.5, 2.4 \pm 0.6, 4.0 \pm 0.9, 2.7 \pm 0.6$ & $73.28\%$ & $91.44\%$ \\
Baseline & $\num{6.43E+09}$ & $3, 2, 4, 3$ & $73.33\%$ & $91.67\%$ \\
\hline
SACT $\tau = 0.005$ & $\num{1.11E+10} \pm \num{4.57E+08}$ & $2.3 \pm 0.4, 3.8 \pm 0.4, 13.1 \pm 2.6, 2.7 \pm 0.5$ & $75.61\%$ & $92.75\%$ \\
Baseline & $\num{1.08E+10}$ & $2, 4, 13, 3$ & $75.57\%$ & $92.86\%$ \\
\hline
SACT $\tau = 0.001$ & $\num{1.44E+10} \pm \num{3.76E+08}$ & $3.0 \pm 0.0, 3.9 \pm 0.2, 19.6 \pm 2.4, 2.9 \pm 0.2$ & $75.84\%$ & $93.09\%$ \\
Baseline & $\num{1.43E+10}$ & $3, 4, 20, 3$ & $76.06\%$ & $93.17\%$ \\
\end{tabular}
}
\caption{Test resolution $224 \times 224$}
\end{subtable}\vspace{5pt}
\begin{subtable}{\linewidth}
\centering
\scalebox{0.9}{
\begin{tabular}{lcccc}
Network & FLOPs & Residual units & Accuracy & Recall@5 \\
\hline
ResNet-50 & $\num{2.02E+10}$ & $3, 4, 6, 3$ & $76.82\%$ & $93.80\%$ \\
ResNet-101 & $\num{3.85E+10}$ & $3, 4, 23, 3$ & $78.37\%$ & $94.60\%$ \\
\hline
ACT $\tau = 0.01$ & $\num{1.58E+10} \pm \num{8.22E+08}$ & $2.9 \pm 0.3, 2.7 \pm 0.5, 3.3 \pm 0.5, 3.0 \pm 0.0$ & $75.82\%$ & $93.18\%$ \\
Baseline & $\num{1.59E+10}$ & $3, 3, 3, 3$ & $75.61\%$ & $93.14\%$ \\
\hline
ACT $\tau = 0.005$ & $\num{2.01E+10} \pm \num{4.19E+08}$ & $3.0 \pm 0.0, 4.0 \pm 0.1, 6.0 \pm 0.4, 3.0 \pm 0.0$ & $76.55\%$ & $93.57\%$ \\
Baseline & $\num{2.02E+10}$ & $3, 4, 6, 3$ & $76.62\%$ & $93.63\%$ \\
\hline
ACT $\tau = 0.001$ & $\num{2.95E+10} \pm \num{2.59E+09}$ & $3.0 \pm 0.0, 4.0 \pm 0.0, 14.6 \pm 2.4, 3.0 \pm 0.0$ & $77.65\%$ & $94.14\%$ \\
Baseline & $\num{2.88E+10}$ & $3, 4, 14, 3$ & $77.73\%$ & $94.31\%$ \\
\hline
ACT $\tau = 0.0005$ & $\num{3.31E+10} \pm \num{2.85E+09}$ & $3.0 \pm 0.0, 4.0 \pm 0.0, 18.0 \pm 2.6, 3.0 \pm 0.0$ & $77.84\%$ & $94.17\%$ \\
Baseline & $\num{3.31E+10}$ & $3, 4, 18, 3$ & $78.10\%$ & $94.43\%$ \\
\hline
SACT $\tau = 0.01$ & $\num{1.65E+10} \pm \num{6.22E+08}$ & $2.6 \pm 0.5, 2.5 \pm 0.6, 4.1 \pm 0.8, 2.8 \pm 0.6$ & $76.34\%$ & $93.43\%$ \\
Baseline & $\num{1.59E+10}$ & $3, 2, 4, 3$ & $75.99\%$ & $93.26\%$ \\
\hline
SACT $\tau = 0.005$ & $\num{2.78E+10} \pm \num{1.13E+09}$ & $2.3 \pm 0.5, 3.9 \pm 0.3, 13.4 \pm 2.7, 2.8 \pm 0.4$ & $78.39\%$ & $94.48\%$ \\
Baseline & $\num{2.67E+10}$ & $2, 4, 13, 3$ & $77.57\%$ & $94.21\%$ \\
\hline
SACT $\tau = 0.001$ & $\num{3.58E+10} \pm \num{9.15E+08}$ & $3.0 \pm 0.0, 4.0 \pm 0.2, 19.9 \pm 2.4, 3.0 \pm 0.2$ & $78.68\%$ & $94.70\%$ \\
Baseline & $\num{3.53E+10}$ & $3, 4, 20, 3$ & $78.23\%$ & $94.38\%$ \\
\end{tabular}
}
\caption{Test resolution $352 \times 352$}
\end{subtable}
\egroup
\caption{ImageNet validation set. Comparison of ResNet, ACT, SACT and the respective baselines. All models are trained with $224 \times 224$ resolution images. $(x \pm y)$ denotes mean value $x$ and one standard deviation $y$.}
\label{table:imagenet-results}
\end{table}

\end{document}